\documentclass[11pt, a4paper, logo, copyright, nonumbering]{main}
\usepackage[authoryear, sort&compress, round]{natbib}
\usepackage{dblfloatfix}
\usepackage{ulem}
\usepackage{caption}
\usepackage{dramatist}
\usepackage{xspace}
\usepackage{pifont} %
\usepackage{tcolorbox}
\tcbuselibrary{breakable} 
\usepackage{xltabular}
\usepackage{longtable}
\usepackage{longtable}
\usepackage{xeCJK}

\usepackage{amsfonts}
\usepackage{amsmath}
\usepackage{amssymb}
\usepackage{lineno}
\usepackage{multirow}
\usepackage{adjustbox}

\usepackage[bottom]{footmisc}

\usepackage{CJKutf8}
\usepackage{setspace}

\usepackage{dsfont}
\usepackage{array} %
\usepackage{subfigure} %
\usepackage{xcolor} %
\usepackage[table]{xcolor}
\usepackage{tabularx}
\usepackage{booktabs}
\usepackage{xspace}

\usepackage{lipsum}  %
\usepackage{multicol} %

\usepackage{epigraph}
\usepackage{soul}
\usepackage{hyperref}
\usepackage{cleveref}

\usepackage{CJKutf8}

\makeatletter
\def\@BTrule[#1]{%
  \ifx\longtable\undefined
    \let\@BTswitch\@BTnormal
  \else\ifx\hline\LT@hline
    \nobreak
    \let\@BTswitch\@BLTrule
  \else
    \let\@BTswitch\@BTnormal
  \fi\fi
  \global\@thisrulewidth=#1\relax
  \ifnum\@thisruleclass=\tw@\vskip\@aboverulesep\else
  \ifnum\@lastruleclass=\z@\vskip\@aboverulesep\else
  \ifnum\@lastruleclass=\@ne\vskip\doublerulesep\fi\fi\fi
  \@BTswitch
}
\makeatother

\addto\extrasenglish{
}

{\begin{list}{}%
        {\setlength{\leftmargin}{#1}}%
        \item[]%
}
{\end{list}}

\reportnumber{001} %

\renewcommand{\today}{}

\title{\centering \modelname: An Embodied-Native Vision-Language-Action Model towards Physical AI}

\def\huggingface{\raisebox{-1.5pt}{\includegraphics[height=1.05em]{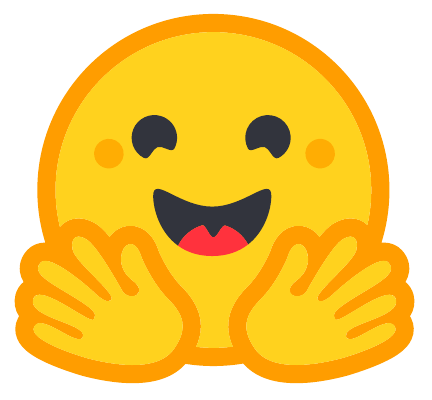}}}
\def\github{\raisebox{-1.5pt}{\includegraphics[height=1.05em]{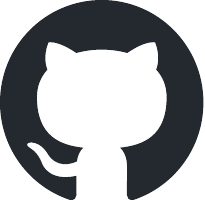}}}

\author[*]{
\textbf{DM0 Team, Dexmal \& StepFun}
\\
\vspace{-0.7em}
\small \github~\textbf{Github}: \url{https://github.com/Dexmal/dexbotic} \\
\small \huggingface~\textbf{Huggingface}: \url{https://huggingface.co/collections/Dexmal/dm0}
}

\renewcommand{\phi}{\varphi}

\renewcommand{\epsilon}{\varepsilon}
\renewcommand{\imath}{\mathrm{i}}

\definecolor{ahared}{RGB}{209, 52, 56}

\newlength{\restsubwidth}
\newlength{\restsubheight}
\newlength{\restsubmoreheight}
\newcommand{\rest}[2]{%
        \settowidth{\restsubwidth}{\ensuremath{#2}}
        \settoheight{\restsubheight}{\ensuremath{{}_{#2}}}
        \ensuremath{{#1\hskip 0.5pt}_{\vrule\kern2pt\parbox[b][%
        4pt][b]{\the\restsubwidth}{%
                        \ensuremath{{}_{#2}}}}}
        }

\newcommand{\modelname}{\textsc{DM0}}

\begin{abstract}

\end{abstract}

\begin{document}

\maketitle

\begin{minipage}{\textwidth}    
   Moving beyond the traditional paradigm of adapting internet-pretrained models to physical tasks, we present \textbf{\modelname}, an Embodied-Native Vision-Language-Action (VLA) framework designed for Physical AI. Unlike approaches that treat physical grounding as a fine-tuning afterthought, \textbf{\modelname} unifies embodied manipulation and navigation by learning from heterogeneous data sources from the onset. Our methodology follows a comprehensive three-stage pipeline: Pretraining, Mid-Training, and Post-Training. First, we conduct large-scale unified pretraining on the Vision-Language Model (VLM) using diverse corpora—seamlessly integrating web text, autonomous driving scenarios, and embodied interaction logs—to jointly acquire semantic knowledge and physical priors. Subsequently, we build a flow-matching action expert atop the VLM. To reconcile high-level reasoning with low-level control, \textbf{\modelname} employs a hybrid training strategy: for embodied data, gradients from the action expert are not backpropagated to the VLM to preserve generalized representations, while the VLM remains trainable on non-embodied data. Furthermore, we introduce an Embodied Spatial Scaffolding strategy to construct spatial Chain-of-Thought (CoT) reasoning, effectively constraining the action solution space. Experiments on the RoboChallenge benchmark demonstrate that \textbf{\modelname} achieves state-of-the-art performance in both Specialist and Generalist settings on Table30.
    
    \begin{center}
        \includegraphics[width=0.8\textwidth]{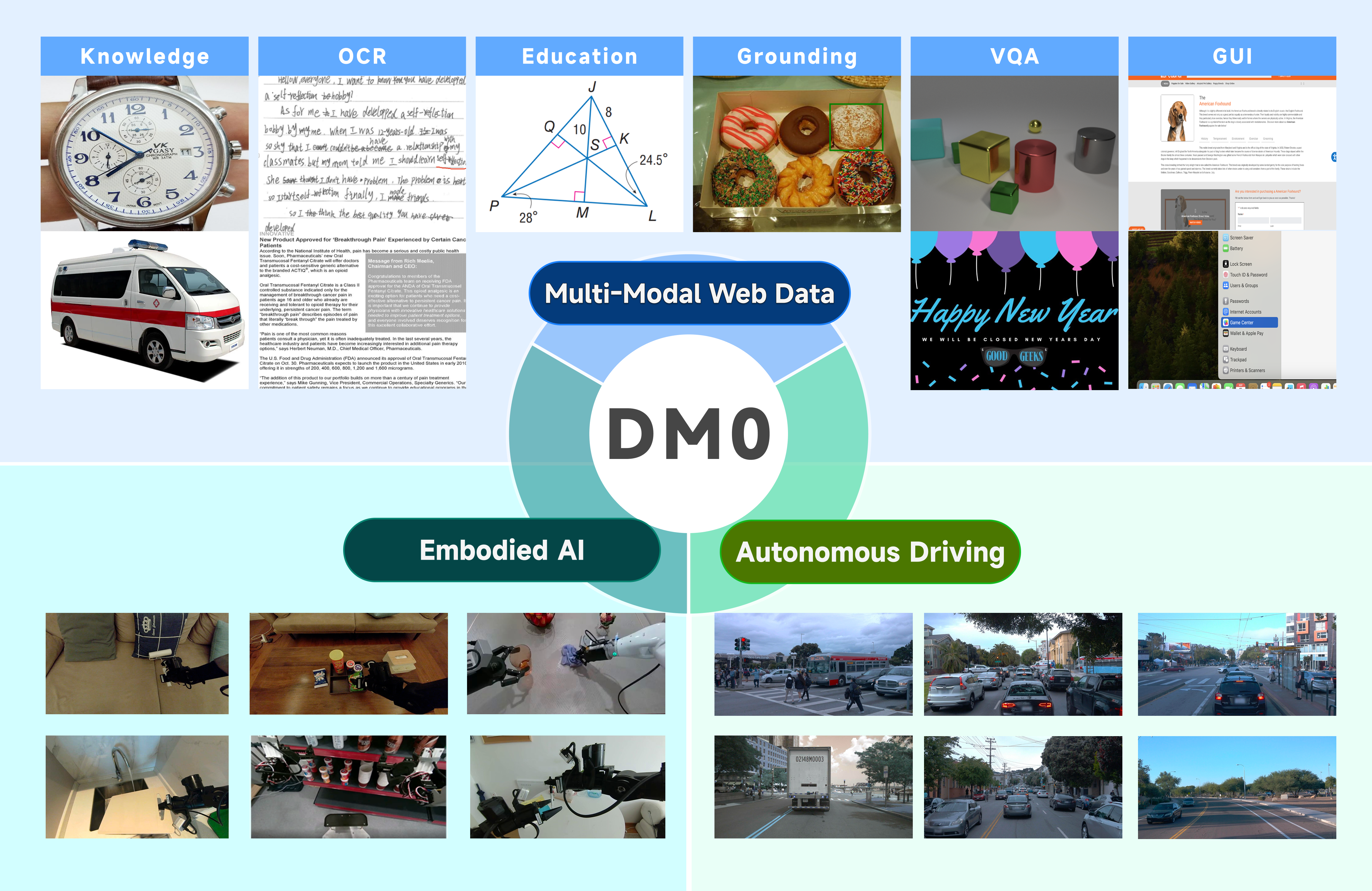}
        \captionof{figure}{\textbf{\modelname} is pretrained on diverse corpora—seamlessly integrating web, autonomous driving and embodied data. It jointly acquire semantic knowledge and physical priors.}
        \label{fig:dm0}
    \end{center}
    
\end{minipage}


\definecolor{colorfirst}{RGB}{252,141,89}
\definecolor{colorsecond}{RGB}{253,187,132}
\definecolor{colorthird}{RGB}{253,212,158}
\definecolor{colorfourth}{RGB}{254,232,200}
\definecolor{colorfifth}{RGB}{255,247,236}
\definecolor{myred}{RGB}{242,128,128}
\definecolor{mygreen}{RGB}{112,180,143}
\definecolor{myblue}{RGB}{210,225,255}
\definecolor{citypink}{RGB}{227,108,194}
\definecolor{cityblue}{RGB}{128,159,225}
\definecolor{casepromptlightblue}{HTML}{66a3ff}
\definecolor{caseanswerdarkblue}{HTML}{025ff7}
\newcommand{\rankfirst}[0]{\cellcolor{colorfirst}}
\newcommand{\ranksecond}[0]{\cellcolor{colorsecond}}
\newcommand{\rankthird}[0]{\cellcolor{colorthird}}
\newcommand{\rankfourth}[0]{\cellcolor{colorfourth}}
\newcommand{\rankfifth}[0]{\cellcolor{colorfifth}}
\DeclareRobustCommand{\legendsquare}[1]{%
  \textcolor{#1}{\rule{2ex}{2ex}}%
}
\DeclareRobustCommand{\legendsquarebox}[1]{%
  \tikz[] \draw[black, fill=#1, line width=0.4pt] (0,0) rectangle (1.5ex,1.5ex);%
}
\newcommand{\cmark}{\textcolor{mygreen}{\ding{51}}}%
\newcommand{\xmark}{\textcolor{myred}{\ding{55}}}%

\definecolor{casebglight}{RGB}{245, 247, 250}
\definecolor{caseborderblue}{RGB}{64, 158, 255}

\newtcolorbox{caseprompt}{
    colback=casebglight,
    colframe=casepromptlightblue,
    title=\textbf{User Prompt},
    fonttitle=\bfseries,
    boxrule=0.5mm,
    sharp corners,
    left=2mm, right=2mm, top=2mm, bottom=2mm
}

\newtcolorbox{casethinking}{
    colback=white,
    colframe=gray!50,
    title=\textbf{Model Thinking Process},
    fonttitle=\bfseries,
    boxrule=0.3mm,
    sharp corners,
    left=2mm, right=2mm, top=2mm, bottom=2mm,
    coltitle=black,
    attach title to upper={\par\vspace{2mm}},
    breakable
}

\newtcolorbox{caseanswer}{
    colback=casebglight,
    colframe=caseanswerdarkblue,
    title=\textbf{Final Answer},
    fonttitle=\bfseries,
    boxrule=0.5mm,
    sharp corners,
    left=2mm, right=2mm, top=2mm, bottom=2mm
}

\newpage

\begin{spacing}{0.9}
\tableofcontents
\end{spacing}

\newpage

\section{Introduction}

The pursuit of Physical AI—artificial intelligence capable of perceiving, reasoning, and acting in the physical world—is a central goal of robotics and computer vision. Recent breakthroughs in Large Language Models (LLMs)~\citep{touvron2023llama,qwen3technicalreport2025,deepseekai2024deepseekv3technicalreport} and Vision-Language Models (VLMs)~\citep{gemmateam2025gemma3technicalreport,huang2026step3vl10btechnicalreport,Qwen3-VL} have endowed agents with impressive semantic understanding and reasoning capabilities. Building upon these foundations, Vision-Language-Action (VLA)~\citep{zitkovich2023rt-2,shi2025memoryvla,sun2025geovla,black2025pi05,black2024pi_0,li2024cogact} models have emerged as a promising paradigm to bridge the gap between high-level cognition and low-level motor control, driving progress towards generalist robots that can operate in unstructured human environments.

However, current VLA research predominantly follows a "Pretrain-then-Adapt" paradigm  ~\citep{xie2025dexbotic,black2025pi05,gr00tn1_2025,wen2025llada}. Typically, models are first pretrained solely on massive static internet datasets (e.g., image-text pairs) and subsequently fine-tuned on limited embodied data. While effective for semantic alignment, this approach suffers from a critical limitation: the model lacks intrinsic physical grounding~\citep{li2024cogact,kim24openvla}. Internet data provides semantic knowledge but fails to capture the dynamic, continuous, and spatial nature of physical interactions. Consequently, adapting these "Internet-Native" models often leads to distinct module fragmentation (separating navigation from manipulation) or catastrophic forgetting, where the pursuit of motor skills degrades the model's general reasoning capacity.

We argue that true generalist robots require an Embodied-Native framework. This entails training the model from scratch (or from very early stages) with a unified perspective that treats embodied sensorimotor data as a first-class citizen alongside linguistic and visual data. Such a framework must harmonize heterogeneous data sources—spanning web corpus, autonomous driving logs, and robot manipulation trajectories—to learn representations that are simultaneously semantically rich and physically actionable.

To realize this vision, we introduce \textbf{\modelname}, an Embodied-Native VLA framework designed to unify manipulation and navigation. Unlike conventional adaptation methods, \textbf{\modelname} is built upon a multi-source, three-stage training pipeline: Pretraining, Mid-Training, and Post-Training. Our framework is anchored by three key components.
\begin{itemize}
    \item \textbf{Unified Pre-training on Vision-Language, Driving and Embodied Corpus:} In the initial stage, we perform large-scale unified pretraining on the VLM using diverse corpora—integrating vision-language data, driving scenes data, and embodied action data. This ensures the model acquires physical priors (e.g., spatial relations, physics dynamics) concurrently with semantic knowledge. 

    \item \textbf{Hybrid Training Architecture:} To translate this understanding into precise actions, we construct a flow-matching action expert upon the VLM. During Mid- and Post-Training, we employ a hybrid gradient strategy: gradients from the action expert are decoupled from the VLM for embodied tasks to prevent the erosion of general knowledge, while the VLM continues to learn from non-embodied data.
    
    \item \textbf{Embodied Spatial Scaffolding Strategy:} To further bridge the reasoning-action gap, we propose an Embodied Spatial Scaffolding strategy. This mechanism generates spatial Chain-of-Thought (CoT) reasoning to decompose complex instructions, effectively constraining the action solution space for the policy.
    
    
\end{itemize}

Extensive experiments on the RoboChallenge benchmark show that \textbf{\modelname} outperforms existing policies, including GigaBrain-0.1~\citep{gigaai2025gigabrain0}, Spirit-v1.5~\citep{spiritai2026spiritv15} and $\pi_{0.5}$~\citep{black2025pi05}, achieving state-of-the-art results on Table30 in both Specialist and Generalist settings. Specifically, \textbf{\modelname}-Specialist achieves an average 62.0\% success rate for thirty Table30 tasks, surpassing previous SoTA GigaBrain-0.1 by more than 10\% margins. \textbf{\modelname}-Generalist produces an average 37.3\% success rate, surpassing $\pi_{0.5}$-Generalist by a large margin, showing superior generalization ability.

\setlength{\epigraphwidth}{0.95\columnwidth}
\renewcommand{\epigraphflush}{center}
\renewcommand{\textflush}{flushepinormal}
\renewcommand{\epigraphsize}{\footnotesize}

\section{Model Description}
\label{sec:model_descrip}


\subsection{Model Architecture}

\begin{figure}[htbp]
  \centering
  \includegraphics[width=\textwidth]{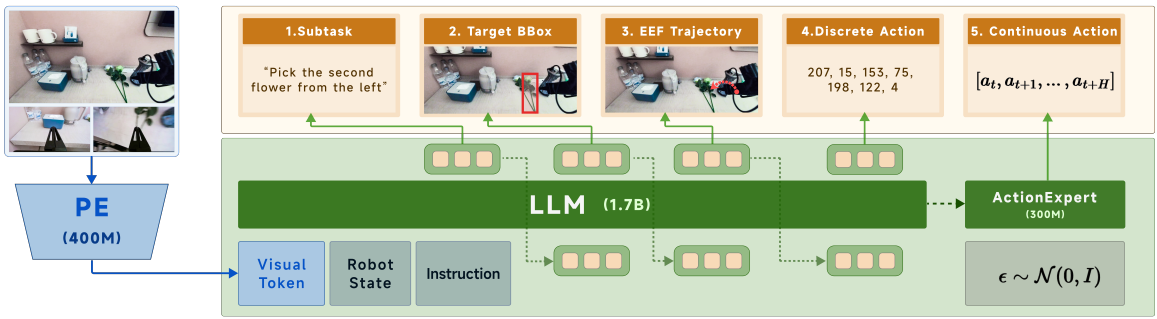}
  \captionsetup{justification=raggedright,singlelinecheck=false}
  \caption{Model Architecture. \textbf{\modelname}~consists of a vision-language model (VLM) backbone and a Flow Matching \citep{lipman2022flowmatching} based action expert. The VLM processes multi-modal inputs and generates embodied reasoning representations, which are subsequently consumed by the action expert to produce continuous robot actions.}
\label{fig:model_architecture}
\end{figure}

The \textbf{\modelname}~model, illustrated in Figure~\ref{fig:model_architecture}, is an end-to-end Vision-Language-Action (VLA) model that supports joint training on large-scale datasets spanning diverse tasks and data distributions, including web-scale multi-modal data, driving-scene data, and embodied data.

The proposed architecture is composed of two core components:
(i) a VLM built on the Qwen3-1.7B large language model (LLM) \citep{qwen3technicalreport2025}, augmented with a perception encoder (PE) \citep{bolya2025perceptionencoder} to enable multi-modal perception, semantic understanding, and embodied reasoning in robotic environments; and
(ii) an action expert based on Flow Matching \citep{lipman2022flowmatching}, which generates continuous control actions conditioned on the key-value (KV) cache extracted from the VLM backbone. The multi-view images are resized to $728\times728$ and input to the PE, after which the image embeddings are downsampled by 4$\times$ using two $3\times3$ convolutional layers with stride 2.

At inference time, \textbf{\modelname}~supports two alternative inference modes.
In the first mode, the model directly predicts continuous action sequences from multi-modal observations and language instructions.
In the second mode, the model first generates textual outputs for embodied reasoning, and subsequently conditions the action expert on these outputs to produce continuous actions.
Formally, the distribution of the joint model is factorized as follows.

\begin{equation}
\pi_\theta 
\left(
  \hat{l}, \mathbf{a}_{t:t+H} \mid \mathbf{o}_t, l
\right) 
= 
\pi_\theta
\left(
  \hat{l} \mid \mathbf{o}_t, l
\right)
\cdot 
\pi_\theta
\left(
  \mathbf{a}_{t:t+H} \mid \mathbf{o}_t, l, \hat{l}
\right).
\end{equation}

Here, $\mathbf{a}_{t:t+H}$ denotes the continuous action sequence over a horizon of $H$ time steps.
The multi-modal observation at time step $t$ is indicated by $\mathbf{o}_t = [\mathbf{I}_t, \mathbf{s}_t]$, where $\mathbf{I}_t$ represents visual inputs and $\mathbf{s}_t$ corresponds to the proprioceptive state of the robot.
The variable $l$ denotes the language instruction and $\hat{l}$ denotes the predicted textual output.

\subsection{Multi-Source Hybrid Training}

Several previous works \citep{driess2025knowledge} \citep{zhang2026clap} have investigated unified training paradigms that integrate VLM with an action expert for end-to-end learning. Although such approaches offer architectural simplicity, the jointly optimization of language and continuous control objectives has been observed to adversely affect the semantic representations preserved in pretrained VLM, potentially undermining their language understanding and reasoning capabilities. 

We adopt a hybrid gradient strategy inspired by Knowledge Insulation (KI) \citep{driess2025knowledge} that decouples gradients from the action expert when training on embodied data, thereby preventing the erosion of semantic knowledge in the pretrained VLM. At the same time, the VLM continues to be updated using non-embodied data, allowing the model to further refine its general language and visual understanding capabilities. In addition, the VLM is supervised to predict discrete action tokens, encouraging it to encode action-relevant semantics that are beneficial for downstream action prediction. We provide a detailed description of the training procedure in the following.

The VLM is trained to predict embodied reasoning text and discrete action tokens by minimizing the auto-regressive cross-entropy loss:

\begin{equation}
\mathcal{L}_{\mathrm{AR}}(\theta)
=
- \mathbb{E}_{\mathcal{D}}
\left[
  \log \pi_{\theta}( \hat{l} \mid \mathbf{o}_t, l)
\right].
\end{equation}

The action expert is trained to predict continuous action sequences by minimizing the Flow Matching loss:

\begin{equation}
\mathcal{L}_{\mathrm{FM}}(\theta)
=
\mathbb{E}_{\mathcal{D}, \epsilon, \tau}
\left\|
  \pi_\theta
  \left(
    \tilde{\mathbf{a}}_{t:t+H}, \mathbf{o}_t, l, \tau
  \right)
  -
  \left( 
    \mathbf{A}_{t:t+H} - \epsilon
  \right)
\right\|^2.
\end{equation}

Where, $\mathbf{A}_{t:t+H}$ denotes the ground-truth continuous action sequence, and $\tilde{\mathbf{a}}_{t:t+H} = \tau\mathbf{A}_{t:t+H} + \left(1-\tau\right)\epsilon$ is the noisy action obtained by injecting Gaussian noise $\epsilon \sim \mathcal{N}\left(\mathbf{0}, \mathbf{I}\right)$. The variable $\tau \in [0,1]$ represents the flow time step.

The overall training objective is defined as a weighted combination of the two losses:

\begin{equation}
\mathcal{L}_{\mathrm{total}}(\theta)
=
\lambda\mathcal{L}_{\mathrm{AR}}(\theta)
+
\mathcal{L}_{\mathrm{FM}}(\theta),
\end{equation}

where $\lambda$ is a scalar weighting coefficient. During joint training, we set $\lambda = 1$.

\subsection{Embodied Spatial Scaffolding}
During the joint training stage, we introduce a set of auxiliary objectives organized as a hierarchical prediction framework to provide structured supervision. Specifically, the model is trained to sequentially perform the following tasks:

\begin{enumerate}[label= (\roman*)]
\item \textbf{Subtask prediction}: Predict a fine-grained task description that decomposes the overall task into a sequence of interpretable and manageable steps.
\item \textbf{Goal bounding box prediction}: Predict the bounding box of the target object or goal-relevant region within the visual observation.
\item \textbf{End-effector trajectory prediction}: Predict the future trajectory of the robot end effector in the primary camera view over a specified horizon.
\item \textbf{Discrete action prediction}: Predict discrete action tokens that represent the robot control commands.
\end{enumerate}

This design induces a natural curriculum over levels of abstraction, guiding the model to progressively transition from high-level semantic reasoning to spatial grounding and ultimately to low-level control.
From a theoretical perspective, this hierarchical supervision introduces a sequence of task-aligned inductive biases that progressively constrain the model's hypothesis space. Each intermediate objective functions as a structured information bottleneck, suppressing task-irrelevant variation while preserving semantically and geometrically meaningful structure. In particular, subtask prediction encourages the model to encode high-level intent, goal bounding box prediction enforces object-centric spatial grounding, and trajectory prediction further aligns the learned representations with action-relevant geometry. Together, these objectives encourage representations that become increasingly structured and causally aligned with the underlying visuomotor decision-making process.


\section{Training Recipe}
\label{sec:train_descrip}

\textbf{\modelname}~is trained in three sequential stages that progress from general vision-language ability to embodied control and finally to deployment-ready policies. We accordingly design three training recipes, each tailored to the objective of its stage. \textbf{Pretraining} establishes a strong multimodal foundation on large-scale web, driving, and embodied data. \textbf{Mid-training} adds action prediction and grounds the model in cross-embodiment robot data while retaining general dialogue ability. \textbf{Post-training} narrows the embodiment and data scope to stabilize visuomotor alignment on a small set of target platforms. We describe each stage below.

\subsection{Pretraining}

The goal of pretraining is to learn a general vision-language model with strong multimodal alignment, fine-grained perception, and broad reasoning over web, document, driving, and embodied data. By jointly training on these heterogeneous sources, the model acquires both semantic knowledge and physical priors (e.g., spatial relations and grounding in real-world scenes). The resulting VLM then serves as the backbone that we extend in mid-training with action prediction and embodiment-specific supervision.

\paragraph{Data construction.}
To support both fine-grained perception and broad reasoning, we build a multimodal pretraining corpus spanning the following domains.

\textbf{Knowledge.}
Web-sourced interleaved data come from Common Crawl, StepCrawl, and keyword-based searches; we filter out pages with an image download failure rate exceeding 90\%, QR-code images, and extreme aspect ratios. Image--text pairs are drawn from open collections (LAION~\citep{schuhmann2022laion}, COYO~\citep{kakaobrain2022coyo-700m}, BLIP-CCS~\citep{li2022blip}, Zero~\citep{xie2023ccmb}) and curated via CLIP-based balanced resampling, keyword retrieval, and caption mining from interleaved context based on CLIP similarity and aesthetic scores. We also construct mosaic inputs by concatenating four images into a single training sample.

\textbf{Education.}
Samples cover K-12 (math, physics, chemistry, geometry, chemical formulas; open and synthetic sources including CoSyn~\citep{yang2025cosyn}), university (STEM, medicine, arts, finance), and adult learning (e.g., driving license, CPA, legal exams). Questions are sourced from licensed examination materials and open problem sets~\citep{sujetfinanceqavision100k, vqamed, pathvqa}; supporting content from textbooks, workbooks, and educational sites.

\textbf{OCR.}
Image-to-text uses real and synthetic images (PaddleOCR~\citep{cui2025paddleocr3}, SynthDog~\citep{kim2022synthdog}). Image-to-code covers Markdown, \LaTeX{}, Matplotlib (open and rule-based synthetic charts), and TikZ/Graphviz. Document-to-text uses document pages (PaddleOCR or MinerU~2.0~\citep{wang24mineru}); document-to-code spans HTML, Markdown, and LaTeX (web/GitHub/arXiv) and open datasets~\citep{doclatex,hme100k,websight,liu2024focus}. An OCR-VQA subset combines open data with QA pairs generated from other OCR tasks.

\textbf{Grounding \& Counting.}
Bounding-box and point-level grounding are derived from open detection datasets (OpenImages~\citep{kuznetsova2020open}, COCO~\citep{lin2014microsoft}, Merlin~\citep{yu2024merlin,yu2025unhackable}, PixMo~\citep{deitke2024pixmo}) and in-house document paragraph detection. Counting data are obtained from public benchmarks~\citep{fcs,locount} and by converting detection annotations into counting supervision.

\textbf{VQA.}
General VQA draws on open benchmarks~\citep{liu2025conflict,zellers2019vcr} and QA pairs generated from captions.

\textbf{GUI.}
The GUI subset~\citep{yan2025step} includes interface captions, knowledge VQA, trajectories (across atomic actions), grounding, and web OCR with element coordinates.

\textbf{Driving.}
Driving-scene samples include depth-aware detection and grounding annotations (category labels, metric depth, bounding boxes in normalized $[0,1000]$ coordinates). License plates are masked where required for privacy compliance.

\textbf{Embodied.}
We include embodied data formulated as grounding and caption QA tasks (e.g., object and region localization, scene and spatial-relation descriptions over robot observations), so the model acquires physical and spatial priors alongside semantic knowledge.

Together, these domains establish the perceptual and reasoning foundation that mid-training subsequently integrates with action supervision.

\paragraph{Training settings.}
Pretraining is conducted in a single stage with all parameters jointly optimized. We optimize with AdamW~\citep{loshchilov2017adamw} ($\beta_1=0.9$, $\beta_2=0.95$, $\epsilon=10^{-8}$, weight decay $0.01$), for 1.2T tokens in 370K steps (global batch 8{,}192, length 4{,}096). The learning rate is scheduled in two phases: 900B tokens with linear decay from $5\times10^{-5}$ to $1\times10^{-5}$ for general representation learning; then 300B tokens with linear decay from $1\times10^{-5}$ to $6\times10^{-6}$ on a higher-quality mix to strengthen OCR, grounding, and reasoning capabilities. The resulting model serves as the backbone for the next stage, where we add action prediction and cross-embodiment control. 

\subsection{Mid-Training}

Building on the pretrained backbone, mid-training introduces action prediction and grounds the model in physical control. This stage uses a single training loop that jointly supervises text tokens, discrete action tokens, and continuous actions via the action expert. As a result, the model learns to couple language reasoning with executable physical actions.

\begin{figure*}[ht]
    \centering
    \includegraphics[width=1.0\textwidth]{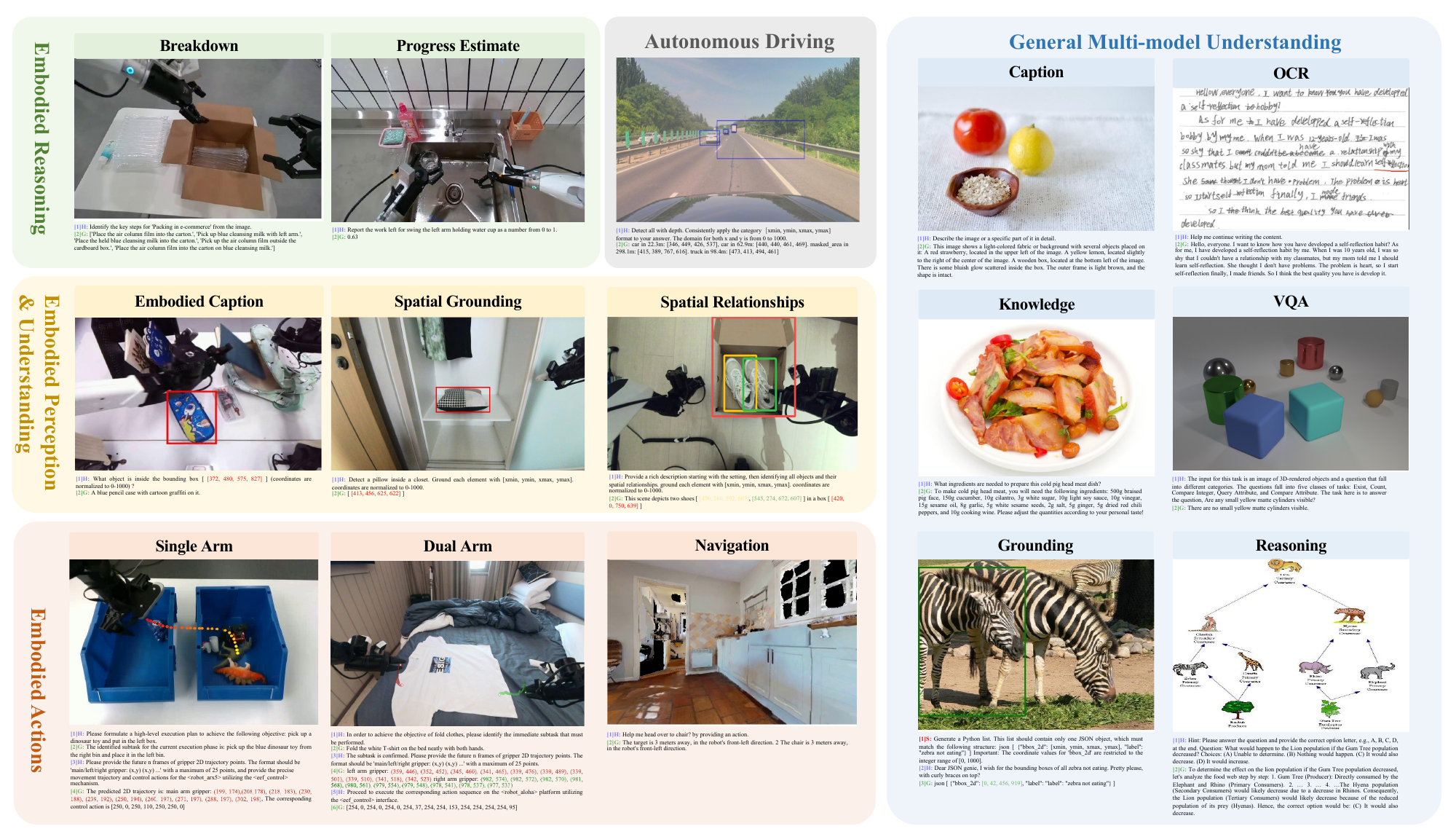}
    \caption{\textbf{Overview of Curated Vision-Language Data.} The curated dataset is designed to enhance core embodied reasoning abilities while preserving the general multimodal understanding and reasoning capabilities of the pretrained VLM.}
    \label{fig:vlm_data}
\end{figure*}

\begin{figure*}[t]
    \centering
    \includegraphics[width=0.9\textwidth]{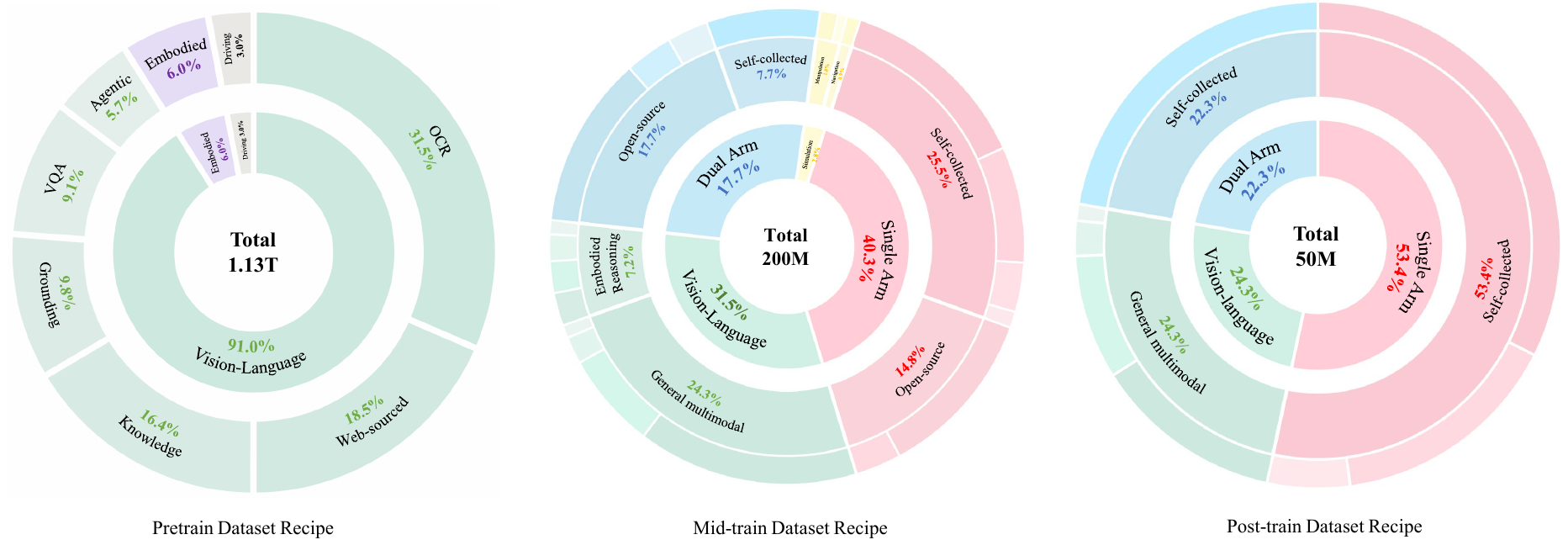}
    \caption{Data mixture ratios for pretraining, midtraining and post-training. After weighted sampling, the total data volumes for the three stages are 1.13T tokens (pretraining), 200M samples (mid-training), and 50M samples (post-training).}
    \label{fig:data_mixture}
\end{figure*}

\paragraph{Goal and supervision.}
Mid-training optimizes three aligned supervision targets: (i) multimodal dialogue tokens, (ii) discretized action trajectories (action tokens), and (iii) continuous action trajectories from the action expert. The latter two correspond to the same underlying sequence, represented in quantized and continuous forms, respectively. Cross-embodiment robot data are mixed with vision-language and reasoning data; we also retain a portion of high-quality vision-language data from pretraining to preserve general multimodal capability.

\paragraph{Data mixture.}
The data mixture (overview in Figure~\ref{fig:vlm_data}, proportions in Figure~\ref{fig:data_mixture}) balances retention of general VLM ability with embodiment-specific learning. It is organized into five categories:
\begin{itemize}
    \item \textbf{Vision--language data.} To preserve general multimodal understanding and instruction-following ability, we include: (1) \texttt{Cambrian-737k}~\citep{tong2024cambrian1}; (2) \texttt{Cambrian-10M (filtered)}~\citep{tong2024cambrian1}, from which we remove low-quality samples and content less relevant to embodied learning (e.g., mathematics-heavy problems, non-English data, purely writing-centric content); (3) \texttt{LLaVA OneVision (OV) 1.5}~\citep{LLaVA-OneVision-1.5}; and (4) \texttt{Self-collected multimodal data}, which includes caption-reannotated embodied data from the pretraining stage (embodied-scene grounding and spatial-relation VQA) as well as other self-collected multimodal data (e.g., GUI grounding and OCR).
    \item \textbf{Embodied reasoning (ER) data.} We construct ER datasets to strengthen high-level planning and temporal reasoning: (1) \textbf{Task Decomposition}, generating the full sequence of steps required to accomplish a task; (2) \textbf{Subtask Prediction}, predicting the next subtask to execute; (3) \textbf{Action QA}, predicting the action that transitions between two frames; (4) \textbf{Temporal reasoning}, recognizing the task being performed and identifying the next subtask; and (5) \textbf{Task Progress Estimation}, estimating the current progress toward task completion.
    \item \textbf{Simulation data.} We include trajectories from (1) four LIBERO~\citep{liu2023libero} tasks (Spatial, Goal, Object, and Long); (2) RoboTwin2.0~\citep{chen2025robotwin} with 50 tasks; and (3) self-collected navigation trajectories in Habitat.
    \item \textbf{Single-arm robotic data.} Trajectories from single-arm manipulation, including (1) self-collected data spanning multiple embodiments (e.g., Franka, UR5, ARX-5, and UMI), and (2) open-source data (OXE~\citep{vuong2023open} and Fuse~\citep{jones2025fuse}).
    \item \textbf{Dual-arm robotic data.} Bi-manual manipulation trajectories, including (1) self-collected data on ALOHA, and (2) open-source data (RoboMind~\citep{wu2025robomind}, Agibot Alpha~\citep{bu2025agibot_arxiv}, and Galaxea Open World~\citep{galaxea2025}).
\end{itemize}

\paragraph{Data representation and processing.}
To support the mixed supervision above, we use two representation styles. Robotic data are stored as episodic JSONL records, where each line corresponds to a timestep with (a) multi-view visual observations (images or video frame references), (b) language instruction, (c) proprioceptive state (used to derive the next-step action by temporal shifting), and optionally \textbf{subtask}, \textbf{goal box}, and \textbf{2D gripper waypoint traces}. Vision--language data are either stored in a ready-to-train conversation format or as minimal keys. In the latter case, full conversations are constructed on-the-fly using a conversation template. From these representations we build (i) multimodal multi-turn dialogues and (ii) continuous trajectory data. The multi-turn dialogue is assembled into a chain-of-thought (CoT) sequence, while the model ultimately predicts the corresponding action trajectory.

\begin{itemize}
    \item \textbf{Trajectory representation.} For robotic data, we train on per-timestep trajectory samples; for vision--language and ER data, we train with multi-turn conversations. Multi-view images are deterministically ordered and truncated or padded to a fixed number of views (three in our setting). If available, 2D waypoint traces and goal boxes are attached as auxiliary supervision.
    \item \textbf{Duplicate data removal.} To reduce redundancy in embodiment data, particularly for end-effector (EEF) trajectories, we apply keyframe sampling. This process involves selecting critical frames that capture significant state changes, thereby filtering out repetitive or static segments before training.
\end{itemize}

\paragraph{Conversation augmentation.}
All data are expressed as template-based conversations so that the same supervision (text, actions, waypoints, etc.) can be presented in varied natural language.
\begin{itemize}
    \item \textbf{Templates and supervision fields.} For robot trajectories, the dialogue is constructed from a task instruction and may optionally include additional supervision fields: \textbf{Subtask} (natural language description), \textbf{2D waypoint trace} (projected gripper waypoints), \textbf{Goal box} (target bounding box), and \textbf{Action} (both discretized action tokens for the VLM and continuous action targets for the action expert). If a supervision field is absent, we automatically select a compatible template that omits that field. For ER and navigation data, we use specialized template families tailored to the supervision type (e.g., QA, reasoning, progress estimation).
    \item \textbf{Template augmentation.} To improve generalization, we design 500 distinct conversation templates for each specific data combination scenario, which are manually polished for quality. During training, we randomly select one of these templates for each sample, introducing linguistic diversity and preventing overfitting to specific prompt structures.
\end{itemize}

\paragraph{Action formulation.}
Actions are supervised in two aligned views over the same sequence: discrete tokens for the VLM and continuous values for the action expert. Per-timestep next actions are constructed via temporal state shifting. We construct short-horizon windows of length 50, normalize them, and quantize them into a 255-bin vocabulary as special action tokens. The VLM predicts tokenized trajectories, while the action expert regresses continuous trajectories.

\paragraph{Training settings.}
Mid-training runs for one epoch on 64$\times$~NVIDIA H20 GPUs with AdamW (learning rate decaying from $2.5\times 10^{-5}$ to $1\times 10^{-5}$), a maximum sequence length of 4{,}096, and AMP enabled. Each sample uses three images with ColorJitter augmentation; per-device batch size is 6. The resulting model supports multimodal conversation and action trajectory prediction across diverse embodiments. Post-training then narrows the embodiment and data scope for deployment.

\subsection{Post-Training}

Post-training starts from the mid-trained model and specializes it for deployment by concentrating robot data on a small set of target embodiments. Narrowing embodiment diversity reduces distributional variance and stabilizes cross-modal alignment, enabling the model to learn fine-grained visuomotor correspondences and action semantics where they matter most for the chosen manipulators.

\paragraph{Data strategy and mixture.}
The data mixture consists of (1) a resampled subset of mid-training vision--language data to preserve general dialogue ability, and (2) single-arm and dual-arm robotic data limited to the target embodiments. Data representation, conversation templates, and action formulation remain identical to those used in mid-training.

\paragraph{Training settings.}
Training continues under the same joint supervision (text, discrete action tokens and continuous action expert targets) and the same optimization configuration as mid-training. Only the data sampling and the set of target embodiments are modified. The resulting model specializes to the target platforms while retaining general multimodal ability.
\section{Experimental Evaluations}
\label{sec:evaluations}


\subsection{Experimental Setup}

\paragraph{Benchmark Protocols.}
Given \modelname's focus on physical-world interaction, we evaluate it on the real-world \textbf{RoboChallenge} benchmark [\citep{robochallenge}], which features a comprehensive suite of over 30 long-horizon tabletop manipulation tasks (see \textbf{Table~30}). These tasks require multi-step reasoning, spatial understanding, and precise continuous control, covering domains such as object picking, placement, rearrangement, tool usage, and compositional instruction following.

\paragraph{Training Setup.}
During the supervised fine-tuning (SFT) phase, we employ two distinct training configurations, differentiated by their data sources and evaluation scopes.
\begin{itemize}
\item \textbf{SFT (Specialist):} This configuration is trained exclusively on data from the target task. The model was fine-tuned on 8$\times$H20 GPUs with a batch size of 4 per GPU for 40K to 150K iterations, contingent upon task complexity. We set the action horizon to 50 steps. For tasks involving repetitive sub-goals, progress supervision was applied to enhance learning efficacy and performance.
\item \textbf{SFT (Generalist):} This configuration is trained on data aggregated from all available tasks for the target robot platform and is consequently evaluated on the full suite of tasks for that robot. Training was conducted on 16$\times$H20 GPUs with a batch size of 4 per GPU for 200K iterations, with an action horizon of 50 steps.
\end{itemize}

\paragraph{Comparison Models.}
We benchmark \modelname~against several leading open-source models, including GigaBrain-0.1 \citep{gigaai2025gigabrain0}, Spirit-V1.5 \citep{spiritai2026spiritv15}, 
$\pi_{0.5}$ \citep{black2025pi05}, and $\pi_0$ \citep{black2024pi_0}.

The benchmark evaluates performance using both success rate and a composite task score. In our analysis, we report the success rate for the specialist setting, but provide both metrics for the generalist setting due to the low success rate for all models.

\subsection{RoboChallenge Results}

In Table~\ref{tab:same_size_bmk_sr}, we benchmark the \modelname~specialist model against strong open-source baselines within the 3B--5B parameter range. The results demonstrate that \modelname, despite having only 2B parameters, consistently outperforms all compared models across multiple robot platforms (UR5, Franka, ARX5, ALOHA) and achieves a superior overall success rate of 62.00\%. The performance advantage is particularly pronounced in complex, long-horizon tasks such as arrange fruits in basket, plug in network cable and sweep rubbish, where \modelname~achieves perfect or near-perfect success while other models frequently fail. This indicates that \modelname's design offers significantly more efficient and effective task-specific manipulation capabilities. 
\begin{table}[htbp]
  \centering
  \scriptsize
  \caption{Comparison with state-of-the-art open-source VLA models on Table30 benchmark in RoboChallenge. List in success rate. Tasks with * use the progress supervision. \textbf{100} for highest and \underline{90} for second highest. The data is reported before 20260210.}
  \label{tab:same_size_bmk_sr}
  \setlength{\tabcolsep}{4pt}
  \renewcommand{\arraystretch}{0.95}
  \begin{adjustbox}{width=\textwidth, max totalheight=0.92\textheight}
  \begin{tabular}{llcccc}
  \toprule
  \multicolumn{2}{c}{\multirow{4}{*}{\textbf{Benchmark}}} & \multicolumn{4}{c}{
  \textbf{Model}} \\
  \cmidrule(lr){3-6}
  & & \modelname & Spirit-v1.5 & GigaBrain-0.1 & $\pi_{0.5}$ \\
  & & 2B & 4B & 3B & 3B \\
  \midrule[0.8pt]
  \multirow{4}{*}{\parbox{1.6cm}{\centering\textbf{UR5}}}
& arrange fruits in basket & \textbf{100} & \underline{80} & 60 & 40 \\
& hang toothbrush cup & \textbf{80} & \textbf{80} & 40 & 50 \\
& set the plates & \textbf{100} & 80 & \underline{90} & 80 \\
& shred scrap paper & \textbf{30} & \underline{20} & 0 & 0 \\
& sort books & \textbf{20} & 0 & 0 & 0 \\
& stack color blocks & \textbf{100} & 80 & \textbf{100} & \textbf{100} \\
  \midrule
  \multirow{2}{*}{\parbox{1.6cm}{\centering\textbf{Franka}}}
& move objects into box & \textbf{100} & 80 & \underline{90} & 50 \\
& press three buttons & \textbf{90}$^*$ & \textbf{90} & 40 & 0 \\

  \midrule
  \multirow{11}{*}{\parbox{1.5cm}{\centering\textbf{ARX5}}}
& arrange flowers & \textbf{70} & \underline{50} & 40 & \underline{50} \\
& arrange paper cups & \underline{30} & 0 & \textbf{80} & 0 \\
& fold dishcloth & \textbf{20} & \textbf{20} & 10 & \textbf{20} \\
& open the drawer & \textbf{100} & 70 & \textbf{100} & 40 \\
& place shoes on rack & \textbf{100} & \underline{90} & 50 & \underline{90} \\
& put cup on coaster & \textbf{100} & 90 & \textbf{100} & 90 \\
& search green boxes & \textbf{100} & \underline{90} & 80 & 80 \\
& sort electronic products & 0 & \underline{30} & 0 & \textbf{50} \\
& turn on light switch & \textbf{80} & \textbf{80} & 60 & 40 \\
& water potted plant & \textbf{80}$^*$ & 0 & \underline{60} & 0 \\
& wipe the table & 0 & 0 & 0 & 0 \\
  \midrule
  \multirow{11}{*}{\parbox{1.5cm}{\centering\textbf{ALOHA}}}
& clean dining table & 0 & \underline{30} & \textbf{40} & 10 \\
& make vegetarian sandwich & 0 & 0 & 0 & 0 \\
& plug in network cable & \textbf{80} & 0 & 0 & \underline{20} \\
& pour fries into plate & 40 & \textbf{50} & \textbf{50} & 30 \\
& put opener in drawer & 30 & \textbf{80} & 40 & \textbf{80} \\
& put pen into pencil case & \underline{90} & \underline{90} & \textbf{100} & 80 \\
& scan QR code & 0 & 0 & \underline{10} & \textbf{50} \\
& stack bowls & \textbf{100} & \textbf{100} & \textbf{100} & \textbf{100} \\
& stick tape to box & \underline{40} & 20 & \textbf{60} & 10 \\
& sweep the rubbish & \textbf{80} & \underline{60} & 50 & 20 \\
& turn on faucet & \textbf{100} & 70 & \textbf{100} & \textbf{100} \\
  \midrule
  & \textbf{Overall} & \textbf{62.00} & 51.00 & \underline{51.67} & 42.67 \\
  \bottomrule
  \end{tabular}
  \end{adjustbox}
  \end{table}

In Table~\ref{tab:same_size_bmk_gen_sr_score}, we compare the \modelname-generalist model against the $\pi_{0}$-generalist and $\pi_{0.5}$-generalist models. The \modelname-generalist achieves a combined average success rate and task score of 37.3/49.08, substantially outperforming both $\pi_{0.5}$-generalist (17.67/31.27) and $\pi_0$-generalist (9.0/20.22). This performance edge is consistent across nearly all robot platforms and task categories. Notably, \modelname-generalist excels in tasks requiring precise manipulation and long-horizon reasoning, such as stack color blocks(100/100.0), place shoes on rack(100/98.5), put cup on coaster (100/100.0), and search green boxes(100/100.5), where it often achieves perfect scores while competing models achieve low scores. The results indicate that \modelname's generalist formulation provides superior cross-task adaptability and robust performance.
\begin{table}[htbp]
  \centering
  \scriptsize
  \caption{Comparison with state-of-the-art open-source VLA generalist models on Table30 benchmark in RoboChallenge. List in success rate / score. \textbf{100/100} for highest score. The data is reported before 20260210.}
  \label{tab:same_size_bmk_gen_sr_score}
  \setlength{\tabcolsep}{4pt}
  \renewcommand{\arraystretch}{0.95}
  \begin{adjustbox}{width=0.9\textwidth, max totalheight=0.92\textheight}
  \begin{tabular}{llccccc}
  \toprule
  \multicolumn{2}{c}{\multirow{4}{*}{\textbf{Benchmark}}} & \multicolumn{3}{c}{\textbf{Model}} \\
  \cmidrule(lr){3-5}
  & & \modelname &  $\pi_{0.5}$ &  $\pi_0$ \\
  & & 2B &  3B  & 3B \\
  \midrule[0.8pt]
  \multirow{6}{*}{\parbox{1.6cm}{\centering\textbf{UR5}}}
   & arrange fruits in basket & \textbf{70/87.0} & 0/9.0 & 0/11.5 \\
   & hang toothbrush cup & \textbf{90/95.0} & 50/71.0 & 20/62.0 \\
   & set the plates & 60/62 & 40/49.5 & \textbf{50/69.5} \\
   & shred scrap paper & \textbf{30/45.0} & 20/36.0 & 20/27.0 \\
   & sort books & 0/8.5 & 0/24.0 & \textbf{10/26.5} \\
   & stack color blocks & \textbf{100/100.0} & 10/30.0 & 30/39.0 \\

  \midrule
  \multirow{2}{*}{\parbox{1.6cm}{\centering\textbf{Franka}}}
   & move objects into box & \textbf{50/64.5} & 20/40.0 & 20/44.5 \\
   & press three buttons & 0/0.0 & \textbf{0/4.0} & 0/0.0 \\
  \midrule
  \multirow{11}{*}{\parbox{1.5cm}{\centering\textbf{ARX5}}}
    & arrange flowers & \textbf{20/49.0} & 0/30.5 & 0/13.5 \\
    & arrange paper cups & \textbf{10/54.0} & 0/31.0 & 0/15.0 \\
    & fold dishcloth & \textbf{10/10.5} & 0/0.0 & 0/0.0 \\
    & open the drawer & \textbf{90/95.0} & 50/80.0 & 0/20.0 \\
    & place shoes on rack & \textbf{100/98.5} & 0/20.0 & 0/16.5 \\
    & put cup on coaster & \textbf{100/100.0} & 70/63.0 & 0/0.0 \\
     & search green boxes & \textbf{100/95.5} & 0/3.0 & 0/0.0 \\
    & sort electronic products & 0/18.4 & \textbf{0/22.5} & \textbf{0/22.5} \\
    & turn on light switch & \textbf{70/70.5} & 10/25.0 & 20/29.0 \\
    & water potted plant & \textbf{0/33.5} & 0/0.0 & 0/0.0 \\
    & wipe the table & \textbf{0/47.5} & 10/28.0 & 0/29.0 \\

  \midrule
    \multirow{11}{*}{\parbox{1.5cm}{\centering\textbf{ALOHA}}}

 & clean dining table & 0/12 & \textbf{30/62.0} & 0/25.5 \\
 & make vegetarian sandwich & \textbf{0/15.0} & 0/0.0 & 0/0.0 \\
 & plug in network cable & \textbf{20/45.5} & 0/0.0 & 0/0.0 \\
 & pour fries into plate & \textbf{0/6.0} & 0/0.0 & 0/0.0 \\
 & put opener in drawer & 10/10.0 & \textbf{20/38.0} & 0/0.0 \\
 & put pen into pencil case & 20/40.0 & \textbf{50/63.5} & 0/14.5 \\
 & scan QR code & 0/0 & \textbf{0/7.0} & 0/3.0 \\
 & stack bowls & 70/71 & \textbf{80/83.0} & 40/53.5 \\
 & stick tape to box & 0/14 & \textbf{0/16.0} & 0/0.0 \\
 & sweep the rubbish & \textbf{30/40.0} & 10/46.0 & 0/17.0 \\
 & turn on faucet & \textbf{70/84.5} & 60/56.0 & 60/67.5 \\
  \midrule
  & \textbf{Overall} & \textbf{37.3/49.08} & 17.67/31.27 & 9.0/20.22\\
  \bottomrule
  \end{tabular}
  \end{adjustbox}
  \end{table}

All evaluated checkpoints of specialist and generalist models and inference code are published in \textbf{https://github.com/dexmal/Dexbotic-RoboChallengeInference}.

\subsection{Multimodal Understanding}

We further evaluate the multimodal capabilities of \modelname~(mid-training checkpoint) on a suite of visual question-answering (VQA) samples. The results indicate that \modelname~retains core VQA functionalities, including scene understanding, visual grounding, attribute recognition, and optical character recognition (OCR), as demonstrated in both embodied (Table \ref{tab:enbody}) and lifestyle (Table \ref{tab:life}) domains. As intended, the model is capable of predicting subtasks, detecting target objects, and forecasting trajectories and actions. Its chain-of-thought (CoT) reasoning performance is detailed in Table \ref{tab:cot}. Moreover, we observe that \modelname~generalizes effectively to mobile contexts, indicating promising potential for deployment in mobile agent applications (Table \ref{tab:mobile}).

\begin{table}[ht]
\centering
\begin{tabular}{@{} m{0.35\linewidth}  m{0.6\linewidth} @{}}
\toprule
\textbf{Image} & \textbf{QAs} \\
\midrule




\includegraphics[width=\linewidth]{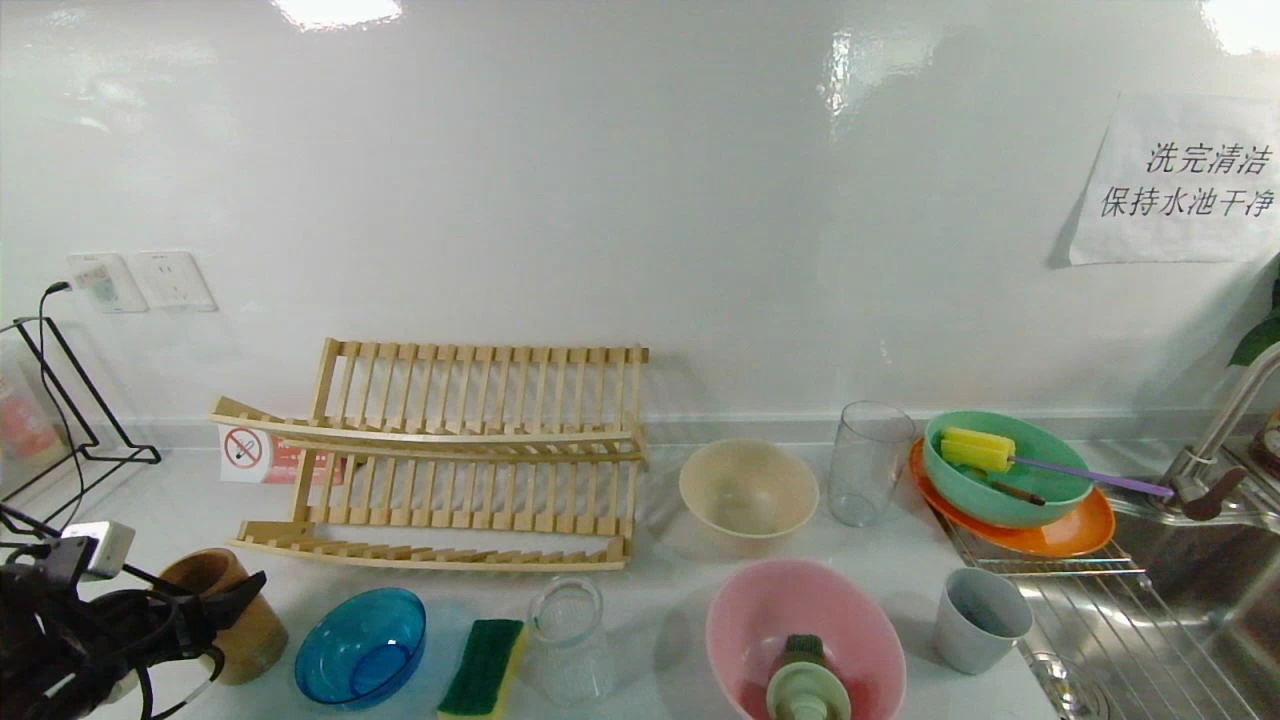} &
{\color{blue}{\textbf{Q:}}} What type is the rack?

{\color{red}{\textbf{A:}}}  wooden rack.

{\color{white}{\textbf{Q:}}}

{\color{blue}{\textbf{Q:}}} How many bowls in the image?

{\color{red}{\textbf{A:}}} 4.

{\color{white}{\textbf{Q:}}}

{\color{blue}{\textbf{Q:}}} What is the object that the robot arm is grasping?

{\color{red}{\textbf{A:}}} Cup.

\\[0.5em]




\midrule
\includegraphics[width=\linewidth]{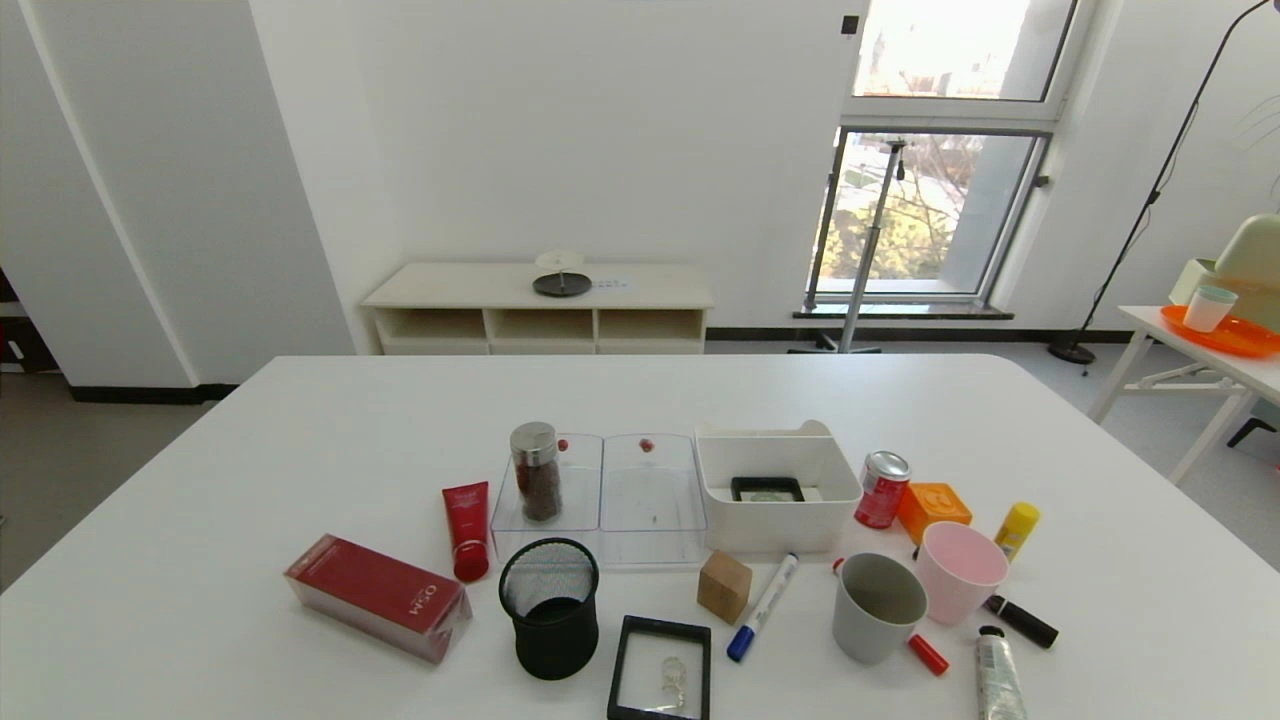} &
{\color{blue}{\textbf{Q:}}} Describe the image.

{\color{red}{\textbf{A:}}}  The scene depicts a white table with various items scattered across its surface. The items include a red toothpaste tube, a red box, a black trash can, a clear plastic container, a white box, a black marker, a blue marker, a pink cup, a yellow marker, a red marker, a white marker, a black eraser, and a small white box.\\
\midrule
\includegraphics[width=\linewidth]{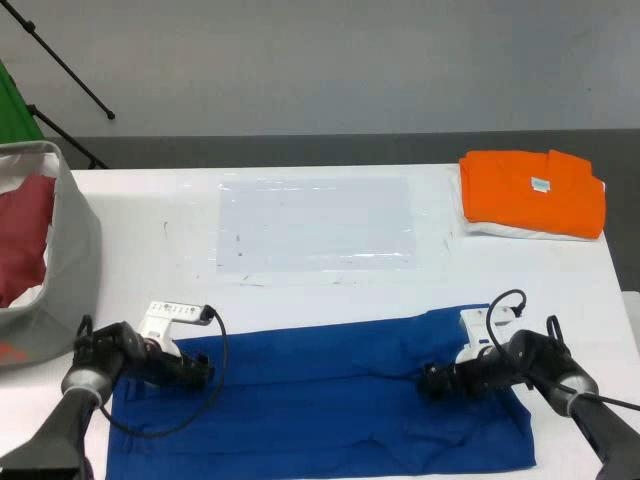}&
{\color{blue}{\textbf{Q:}}}Describe the content of image. 

{\color{red}{\textbf{A:}}}  The image shows a pair of robotic arms positioned on a white table, gripping the handles of a blue fabric. To the left, a gray bag contains red fabric. An orange folded item rests on the table to the right.\\


\midrule
\includegraphics[width=\linewidth]{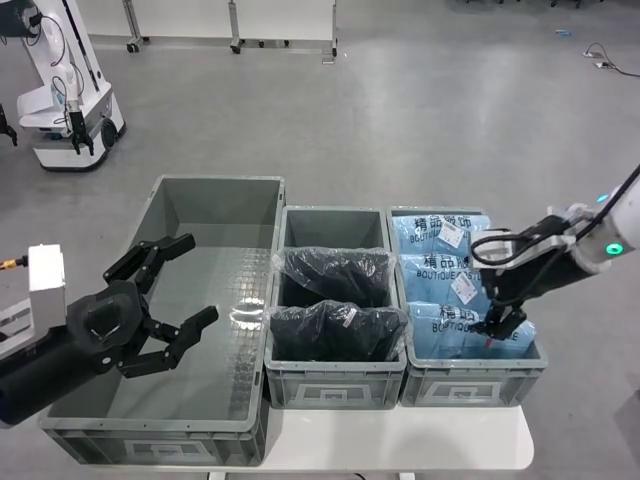} &
{\color{blue}{\textbf{Q:}}} What is the color of clothes in the right box?

{\color{red}{\textbf{A:}}}  clothes in the right box are blue.

{\color{white}{Q:}}

{\color{blue}{\textbf{Q:}}} What is the color of clothes in the middle box?

{\color{red}{\textbf{A:}}}  clothes in the middle box are black.\\
\bottomrule
\end{tabular}
\caption{\modelname~retains fundamental VQA capabilities such as scene description, object detection, and attribute recognition in embodied scenarios. }
\label{tab:enbody}
\end{table}

\begin{table}[ht]
\centering
\begin{tabular}{@{} m{0.35\linewidth}  m{0.6\linewidth} @{}}
\toprule
\textbf{Image} & \textbf{QAs} \\
\midrule
\includegraphics[width=\linewidth]{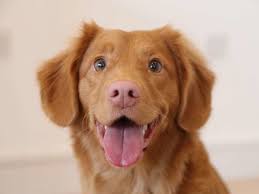} &
{\color{blue}{\textbf{Q:}}} Describe the image.

{\color{red}{\textbf{A:}}} The image shows a dog with a light brown coat, sitting and smiling at the camera. The dog's eyes are open and its mouth is slightly open, revealing a pink tongue. The background is blurred, but it appears to be a indoor setting with a white wall and a white floor. The dog's ears are slightly pointed, and its face is clean and well-groomed. The overall mood of the image is cheerful and friendly.
\\[0.5em]
\midrule
\includegraphics[width=\linewidth]{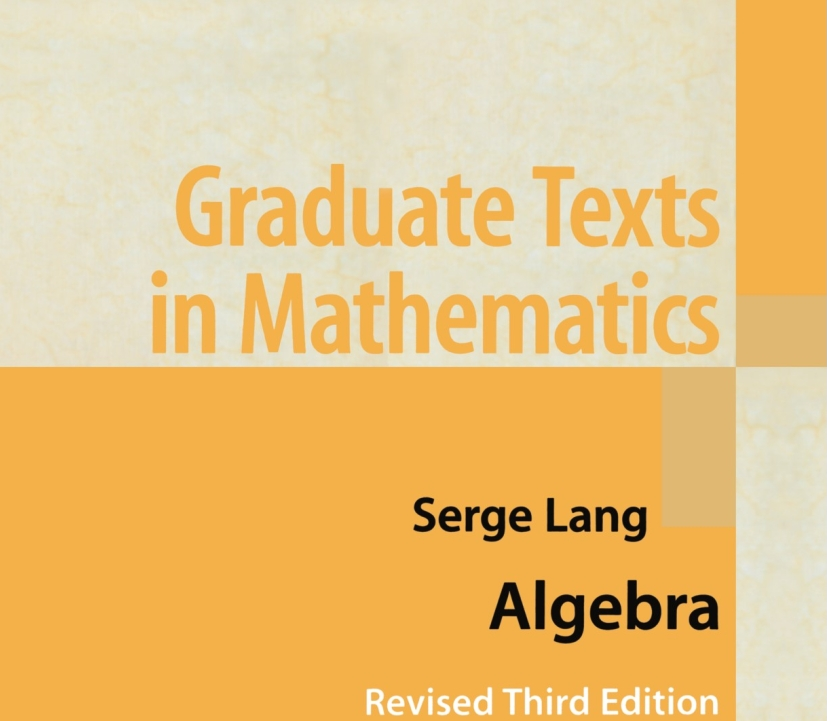} &
{\color{blue}{\textbf{Q:}}} What is the name of this book?

{\color{red}{\textbf{A:}}}  The name of the book is "Algebra" by Serge Lang.

{\color{white}{\textbf{Q:}}}

{\color{blue}{\textbf{Q:}}} What color is the word "Algebra"?

{\color{red}{\textbf{A:}}}  The word "Algebra" is black.
\\[0.5em]
\midrule
\includegraphics[width=\linewidth]{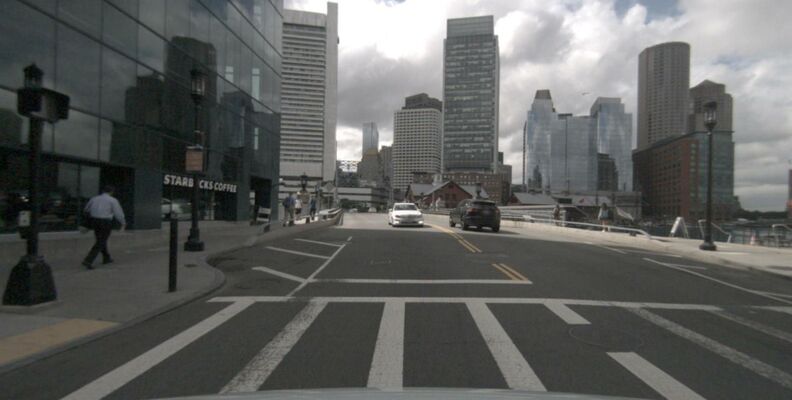} &
{\color{blue}{\textbf{Q:}}} Descibe the image.

{\color{red}{\textbf{A:}}}  The image captures a bustling city street, teeming with life and activity. The street, marked with a crosswalk, serves as the main focus of the image. It's lined with tall buildings, their glass facades reflecting the overcast sky above. The buildings vary in height and design, creating a dynamic skyline that adds depth to the scene. A few cars are visible on the street, their presence adding a sense of movement and life to the image. People can be seen walking on the sidewalk, going about their day, their figures small against the backdrop of the cityscape. The sky above is filled with clouds, casting a soft light over the scene. Despite the overcast conditions, the image exudes a sense of energy and vibrancy, characteristic of a busy city street.
\\
\bottomrule
\end{tabular}
\caption{\modelname~retains fundamental VQA capabilities such as scene description, OCR in general scenarios.}
\label{tab:life}
\end{table}

\begin{table}[ht]
\centering
\begin{tabular}{@{} m{0.35\linewidth}  m{0.6\linewidth} @{}}
\toprule
\textbf{Image} & \textbf{QAs} \\
\midrule


\includegraphics[width=\linewidth]{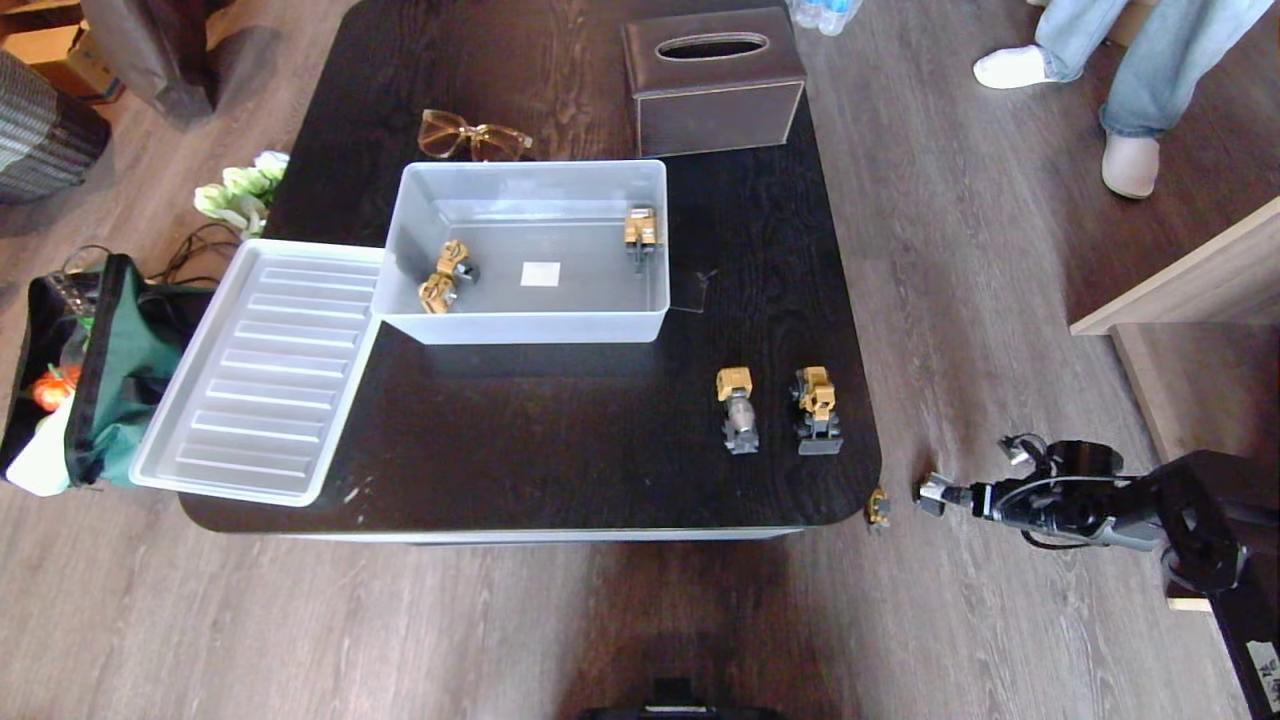} &
{\color{blue}{\textbf{Q:}}} Please decompose the specified high-level goal 'organize toys' into its constituent subtasks.

{\color{red}{\textbf{A:}}}  Pick up the toy car from the ground with your right hand and place it into the storage box.\\
\midrule
\includegraphics[width=\linewidth]{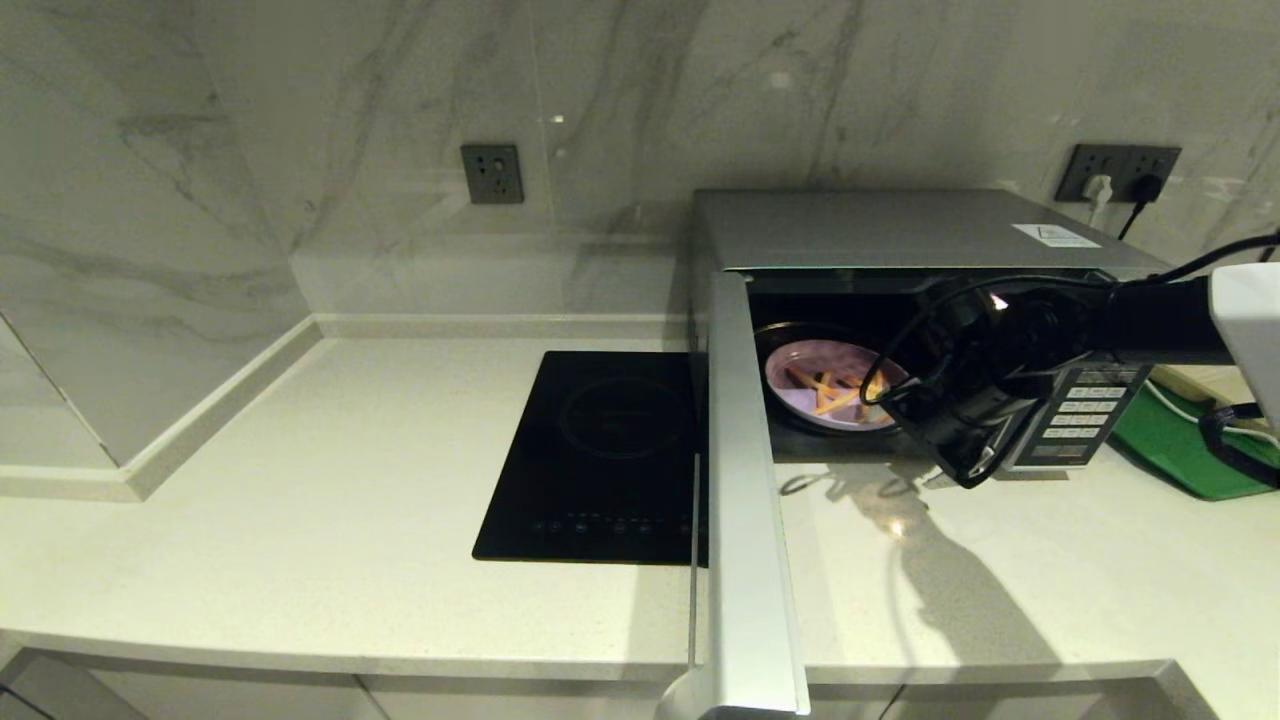} &
{\color{blue}{\textbf{Q:}}} Please identify the initial subtask required to achieve the following objective: open and close microwave oven.

{\color{red}{\textbf{A:}}}  Place the plate from the microwave back on the right side of the table with your right hand.\\
\midrule
\includegraphics[width=\linewidth]{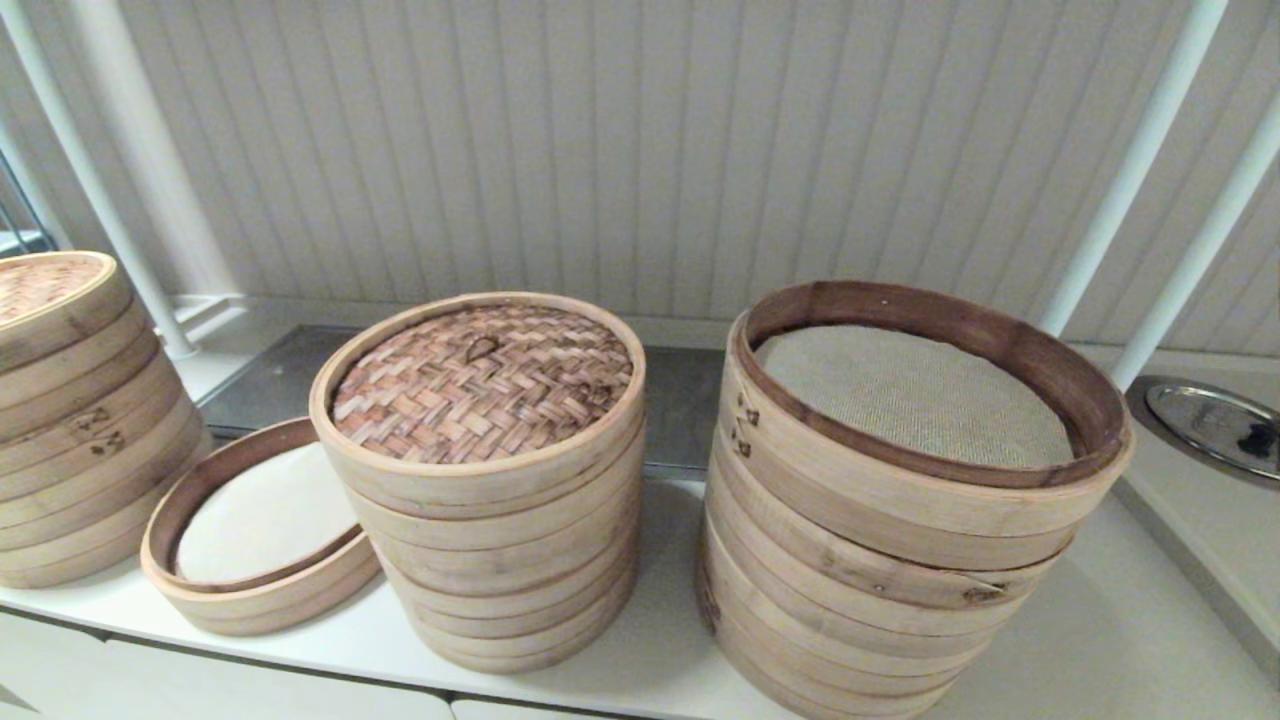} &
{\color{blue}{\textbf{Q:}}} Please identify the specific subtask required to achieve the following goal: lift the steamer lid.

{\color{red}{\textbf{A:}}}  Pick up the second steamer lid from the right with your left hand and place it on the third steamer from the right.\\
\midrule
\includegraphics[width=\linewidth]{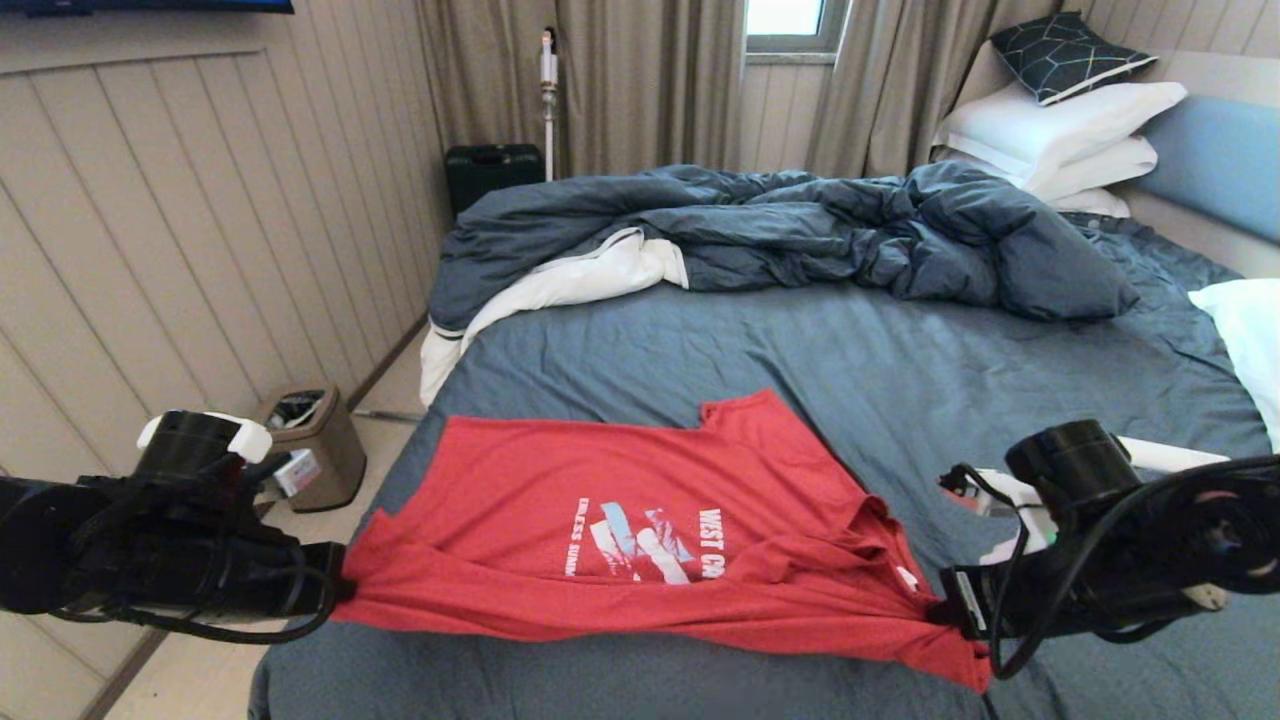} &
{\color{blue}{\textbf{Q:}}} In order to complete the instruction 'fold clothes', please identify the necessary subtask plan.

{\color{red}{\textbf{A:}}}  The current subtask is determined to be: Fold the red T-shirt on the bed neatly with both hands.\\
\bottomrule
\end{tabular}
\caption{\modelname~possesses Chain-of-Thought (CoT) capability for predicting subtasks within embodied scenarios.}
\label{tab:cot}
\end{table}

\begin{table}[ht]
\centering
\begin{tabular}{@{} m{0.1\linewidth}| m{0.4\linewidth}  m{0.4\linewidth} @{}}
\toprule

\textbf{Image} &

\includegraphics[width=\linewidth]{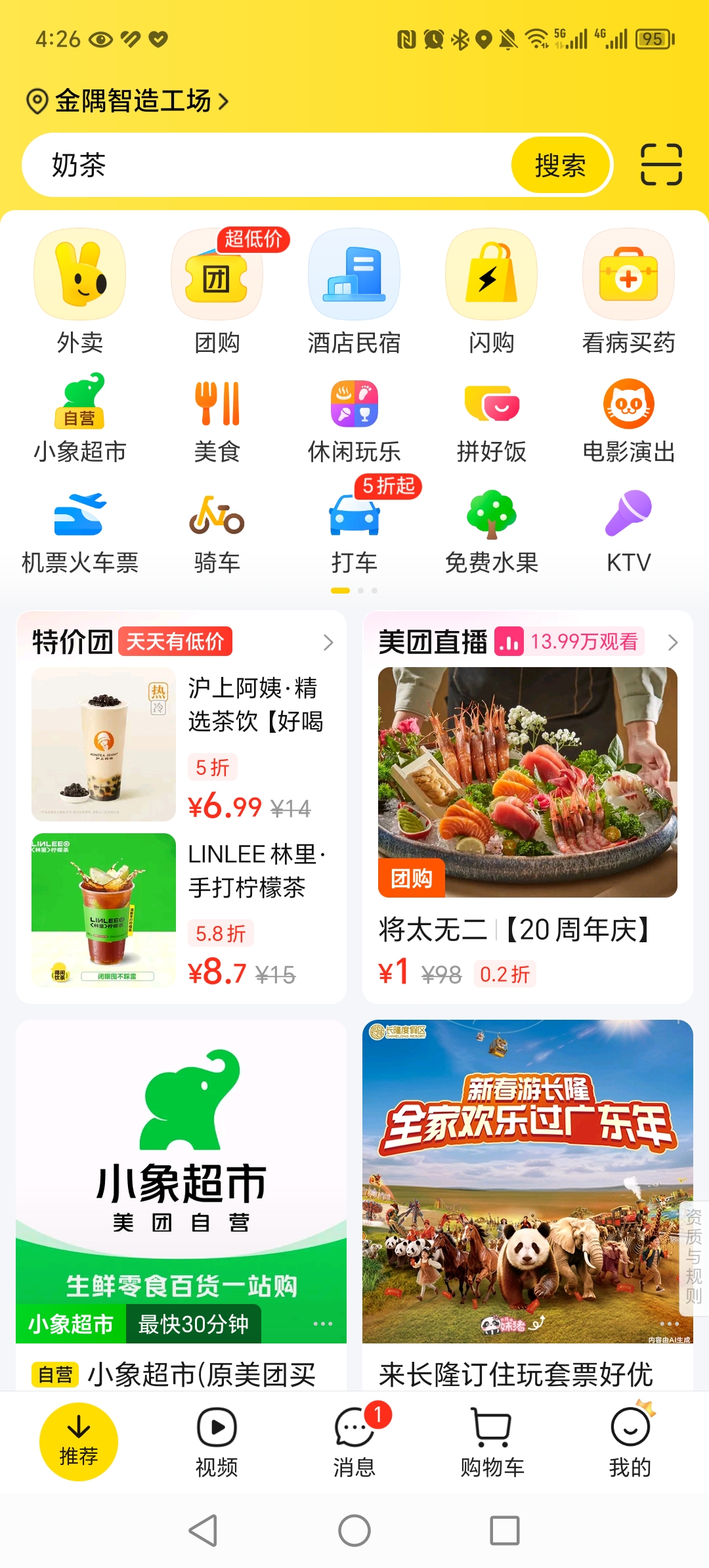} &

\includegraphics[width=\linewidth]{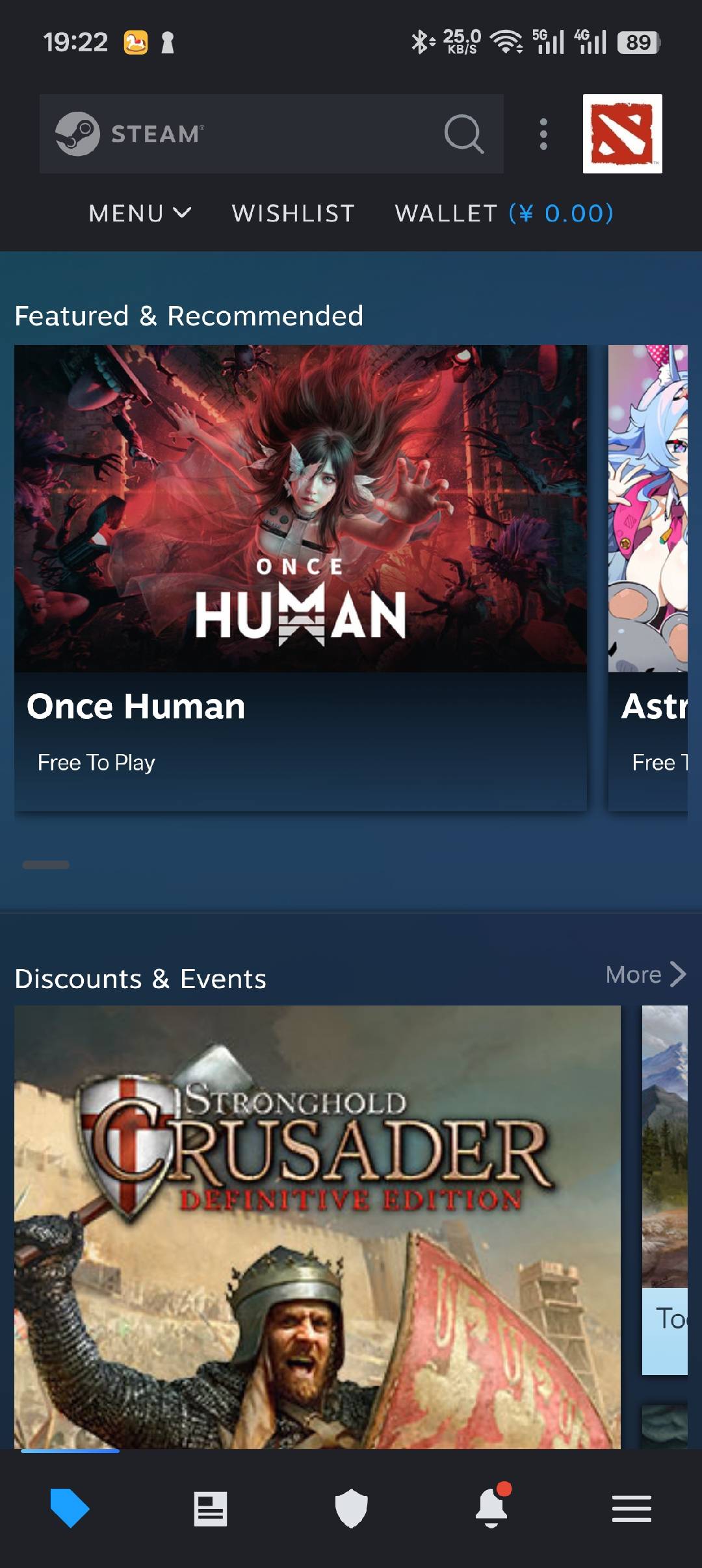} 
\\[0.5em]
\midrule
\textbf{QAs} &
{\color{blue}{\textbf{Q:}}} I want to order my dinner, which button should I press?

{\color{red}{\textbf{A:}}} Press the 外卖 button. &
{\color{blue}{\textbf{Q:}}} I want to check which games are in discounts, which button should I press?

{\color{red}{\textbf{A:}}}  To check which games are in discounts, press the "More" button located at the top right corner of the "Discounts \& Events" section.
\\
\bottomrule
\end{tabular}
\caption{\modelname~also demonstrates potential for supporting mobile agents.}
\label{tab:mobile}
\end{table}
\section{Discussion}
\label{sec:discuss}

\subsection{Conclusion}
\label{conclusion}
In this work, we presented DM0, a pioneering Embodied-Native Vision-Language-Action (VLA) framework that fundamentally rethinks the development of generalist robot policies. Deviating from the conventional paradigm of adapting internet-pretrained models via post-hoc fine-tuning, DM0 integrates physical grounding from the onset through a unified pretraining strategy on heterogeneous data sources—spanning web corpora, autonomous driving logs, and embodied trajectories.We introduced a comprehensive three-stage training pipeline (Pre-, Mid-, and Post-Training) underpinned by two key technical innovations: a hybrid gradient strategy that decouples action expert learning from VLM semantic preservation, and an Embodied Spatial Scaffolding mechanism that leverages spatial Chain-of-Thought (CoT) to rationalize complex physical tasks. Empirical results on the Table30 benchmark demonstrate that DM0 achieves state-of-the-art performance, surpassing strong baselines such as Spirit-v1.5 and $\pi_{0.5}$ in both Specialist and Generalist settings. These findings validate our core hypothesis: constructing VLA models with intrinsic, multi-source physical priors is a more effective path towards robust Physical AI than adapting purely semantic models.

\subsection{Future Work}
While DM0 establishes a strong baseline for Embodied-Native VLAs, several promising directions remain for future exploration:

\begin{itemize}
    \item \textbf{Scaling the Embodied-Native Paradigm:} Currently, DM0 operates as a lightweight model. We plan to investigate the scaling laws of the Embodied-Native framework by training on significantly larger datasets (e.g., combining simulation data with real-world logs) and increasing model parameter size (e.g., to 7B or 30B scales) to observe emergent physical reasoning capabilities.

    \item \textbf{Expanding Multi-Modal Perception:} Physical interaction often requires cues beyond vision and text. Future iterations of DM0 will incorporate additional modalities such as tactile feedback, audio, and depth information directly into the unified pretraining stage, further enhancing the model's fine-grained manipulation skills in occluded or dynamic environments.
    
    \item \textbf{Long-Horizon Reasoning \& World Models:} Although our Spatial Scaffolding improves reasoning, extremely long-horizon tasks remain a challenge. We aim to integrate World Model capabilities into the DM0 framework, allowing the agent to mentally simulate action consequences and plan over longer temporal horizons before execution.

\end{itemize}

\label{future_work}

\newpage
\bibliography{main}

\begin{thebibliography}{58}
\providecommand{\natexlab}[1]{#1}
\providecommand{\url}[1]{\texttt{#1}}
\expandafter\ifx\csname urlstyle\endcsname\relax
  \providecommand{\doi}[1]{doi: #1}\else
  \providecommand{\doi}{doi: \begingroup \urlstyle{rm}\Url}\fi

\bibitem[An et~al.(2025)An, Xie, Yang, Zhang, Zhao, Cheng, Wang, Xu, Chen, Wu, Tan, Li, Yang, Yu, Wang, Qin, Wang, Yan, Feng, Liu, Li, and Deng]{LLaVA-OneVision-1.5}
X.~An, Y.~Xie, K.~Yang, W.~Zhang, X.~Zhao, Z.~Cheng, Y.~Wang, S.~Xu, C.~Chen, C.~Wu, H.~Tan, C.~Li, J.~Yang, J.~Yu, X.~Wang, B.~Qin, Y.~Wang, Z.~Yan, Z.~Feng, Z.~Liu, B.~Li, and J.~Deng.
\newblock Llava-onevision-1.5: Fully open framework for democratized multimodal training.
\newblock In \emph{arXiv}, 2025.

\bibitem[Bai et~al.(2025)Bai, Cai, Chen, Chen, Chen, Cheng, Deng, Ding, Gao, Ge, Ge, Guo, Huang, Huang, Huang, Hui, Jiang, Li, Li, Li, Li, Lin, Lin, Liu, Liu, Liu, Liu, Liu, Liu, Lu, Luo, Lv, Men, Meng, Ren, Ren, Song, Sun, Tang, Tu, Wan, Wang, Wang, Wang, Wang, Xie, Xu, Xu, Xu, Yang, Yang, Yang, Yang, Yu, Zhang, Zhang, Zhang, Zheng, Zhong, Zhou, Zhou, Zhou, Zhu, and Zhu]{Qwen3-VL}
S.~Bai, Y.~Cai, R.~Chen, K.~Chen, X.~Chen, Z.~Cheng, L.~Deng, W.~Ding, C.~Gao, C.~Ge, W.~Ge, Z.~Guo, Q.~Huang, J.~Huang, F.~Huang, B.~Hui, S.~Jiang, Z.~Li, M.~Li, M.~Li, K.~Li, Z.~Lin, J.~Lin, X.~Liu, J.~Liu, C.~Liu, Y.~Liu, D.~Liu, S.~Liu, D.~Lu, R.~Luo, C.~Lv, R.~Men, L.~Meng, X.~Ren, X.~Ren, S.~Song, Y.~Sun, J.~Tang, J.~Tu, J.~Wan, P.~Wang, P.~Wang, Q.~Wang, Y.~Wang, T.~Xie, Y.~Xu, H.~Xu, J.~Xu, Z.~Yang, M.~Yang, J.~Yang, A.~Yang, B.~Yu, F.~Zhang, H.~Zhang, X.~Zhang, B.~Zheng, H.~Zhong, J.~Zhou, F.~Zhou, J.~Zhou, Y.~Zhu, and K.~Zhu.
\newblock Qwen3-vl technical report.
\newblock \emph{arXiv preprint arXiv:2511.21631}, 2025.

\bibitem[{Ben Abacha} et~al.(2019){Ben Abacha}, Hasan, Datla, Liu, Demner-Fushman, and M{\"u}ller]{vqamed}
A.~{Ben Abacha}, S.~A. Hasan, V.~V. Datla, J.~Liu, D.~Demner-Fushman, and H.~M{\"u}ller.
\newblock Vqa-med: Overview of the medical visual question answering task at imageclef 2019.
\newblock In \emph{Working Notes of CLEF 2019}, volume 2380 of \emph{CEUR Workshop Proceedings}, Lugano, Switzerland, 2019.

\bibitem[Black et~al.(2024)Black, Brown, Driess, Esmail, Equi, Finn, Fusai, Groom, Hausman, Ichter, Jakubczak, Jones, Ke, Levine, Li-Bell, Mothukuri, Nair, Pertsch, Shi, Tanner, Vuong, Walling, Wang, and Zhilinsky]{black2024pi_0}
K.~Black, N.~Brown, D.~Driess, A.~Esmail, M.~Equi, C.~Finn, N.~Fusai, L.~Groom, K.~Hausman, B.~Ichter, S.~Jakubczak, T.~Jones, L.~Ke, S.~Levine, A.~Li-Bell, M.~Mothukuri, S.~Nair, K.~Pertsch, L.~X. Shi, J.~Tanner, Q.~Vuong, A.~Walling, H.~Wang, and U.~Zhilinsky.
\newblock $\pi_0$: A vision-language-action flow model for general robot control.
\newblock \emph{arXiv preprint arXiv:2410.24164}, 2024.

\bibitem[Black et~al.(2025)Black, Brown, Darpinian, Dhabalia, Driess, Esmail, Equi, Finn, Fusai, Galliker, et~al.]{black2025pi05}
K.~Black, N.~Brown, J.~Darpinian, K.~Dhabalia, D.~Driess, A.~Esmail, M.~R. Equi, C.~Finn, N.~Fusai, M.~Y. Galliker, et~al.
\newblock $\pi_{0.5}$: a vision-language-action model with open-world generalization.
\newblock In \emph{9th Annual Conference on Robot Learning}, 2025.

\bibitem[Bolya et~al.(2025)Bolya, Huang, Sun, Cho, Madotto, Wei, Ma, Zhi, Rajasegaran, Rasheed, Wang, Monteiro, Xu, Dong, Ravi, Li, Dollár, and Feichtenhofer]{bolya2025perceptionencoder}
D.~Bolya, P.-Y. Huang, P.~Sun, J.~H. Cho, A.~Madotto, C.~Wei, T.~Ma, J.~Zhi, J.~Rajasegaran, H.~Rasheed, J.~Wang, M.~Monteiro, H.~Xu, S.~Dong, N.~Ravi, D.~Li, P.~Dollár, and C.~Feichtenhofer.
\newblock Perception encoder: The best visual embeddings are not at the output of the network.
\newblock \emph{arXiv preprint arXiv:2504.13181}, 2025.
\newblock URL \url{https://arxiv.org/abs/2504.13181}.

\bibitem[Bu et~al.(2025)Bu, Cai, Chen, Cui, Ding, Feng, Gao, He, Huang, Jiang, et~al.]{bu2025agibot_arxiv}
Q.~Bu, J.~Cai, L.~Chen, X.~Cui, Y.~Ding, S.~Feng, S.~Gao, X.~He, X.~Huang, S.~Jiang, et~al.
\newblock Agibot world colosseo: A large-scale manipulation platform for scalable and intelligent embodied systems.
\newblock \emph{arXiv preprint arXiv:2503.06669}, 2025.

\bibitem[Byeon et~al.(2022)Byeon, Park, Kim, Lee, Baek, and Kim]{kakaobrain2022coyo-700m}
M.~Byeon, B.~Park, H.~Kim, S.~Lee, W.~Baek, and S.~Kim.
\newblock Coyo-700m: Image-text pair dataset.
\newblock \url{https://github.com/kakaobrain/coyo-dataset}, 2022.

\bibitem[Chai et~al.(2025)Chai, Shen, Zhang, Zhang, Wang, Dou, Kang, Zhang, and Zhang]{doclatex}
M.~Chai, Z.~Shen, C.~Zhang, Y.~Zhang, X.~Wang, S.~Dou, J.~Kang, J.~Zhang, and Q.~Zhang.
\newblock Docfusion: A unified framework for document parsing tasks.
\newblock \emph{arXiv preprint arXiv:2412.12505}, 2025.

\bibitem[Chen et~al.(2025)Chen, Chen, Chen, Cai, Liu, Li, Liang, Lin, Ge, Gu, et~al.]{chen2025robotwin}
T.~Chen, Z.~Chen, B.~Chen, Z.~Cai, Y.~Liu, Z.~Li, Q.~Liang, X.~Lin, Y.~Ge, Z.~Gu, et~al.
\newblock Robotwin 2.0: A scalable data generator and benchmark with strong domain randomization for robust bimanual robotic manipulation.
\newblock \emph{arXiv preprint arXiv:2506.18088}, 2025.

\bibitem[Cui et~al.(2025)Cui, Sun, Lin, Gao, Zhang, Liu, Wang, Zhang, Zhou, Liu, et~al.]{cui2025paddleocr3}
C.~Cui, T.~Sun, M.~Lin, T.~Gao, Y.~Zhang, J.~Liu, X.~Wang, Z.~Zhang, C.~Zhou, H.~Liu, et~al.
\newblock Paddleocr 3.0 technical report.
\newblock \emph{arXiv preprint arXiv:2507.05595}, 2025.

\bibitem[DeepSeek-AI(2024)]{deepseekai2024deepseekv3technicalreport}
DeepSeek-AI.
\newblock Deepseek-v3 technical report, 2024.
\newblock URL \url{https://arxiv.org/abs/2412.19437}.

\bibitem[Deitke et~al.(2024)Deitke, Clark, Lee, Tripathi, Yang, Park, Salehi, Muennighoff, Lo, Soldaini, et~al.]{deitke2024pixmo}
M.~Deitke, C.~Clark, S.~Lee, A.~Tripathi, Y.~Yang, J.~Park, M.~Salehi, N.~Muennighoff, K.~Lo, L.~Soldaini, et~al.
\newblock Molmo and pixmo: Open weights and open data for state-of-the-art vision-language models.
\newblock \emph{arXiv preprint arXiv:2409.17146}, 2024.

\bibitem[Driess et~al.(2025)Driess, Springenberg, Ichter, Yu, Li-Bell, Pertsch, Ren, Walke, Vuong, Shi, and Levine]{driess2025knowledge}
D.~Driess, J.~T. Springenberg, B.~Ichter, L.~Yu, A.~Li-Bell, K.~Pertsch, A.~Z. Ren, H.~Walke, Q.~Vuong, L.~X. Shi, and S.~Levine.
\newblock Knowledge insulating vision-language-action models: Train fast, run fast, generalize better.
\newblock \emph{arXiv preprint arXiv:2505.23705}, 2025.

\bibitem[GigaAI(2025)]{gigaai2025gigabrain0}
GigaAI.
\newblock Gigabrain-0: A world model-powered vision-language-action model.
\newblock 2025.
\newblock URL \url{https://arxiv.org/abs/2510.19430}.

\bibitem[He et~al.(2020)He, Zhang, Mou, Xing, and Xie]{pathvqa}
X.~He, Y.~Zhang, L.~Mou, E.~Xing, and P.~Xie.
\newblock Pathvqa: 30000+ questions for medical visual question answering.
\newblock \emph{arXiv preprint arXiv:2003.10286}, 2020.

\bibitem[Huang et~al.(2026)Huang, Yao, Han, Wan, Guo, Lv, Zhou, Wang, Zhou, Sun, Hu, Lin, Zhao, Huang, Yuan, Qu, Wang, Lai, Zhao, Zhang, Shi, Chen, Weng, Meng, Li, Kong, Dong, Wan, Wang, Qi, Li, Yu, Li, Yin, Zhou, Zhang, Yan, Zhou, Peng, Zhang, Lv, Fu, Cheng, Zhou, Yin, Xie, Wu, Zhang, Liu, Tan, Yan, Chen, Chen, Li, Zhao, Sun, Pang, Fan, Shang, Zhang, You, Ji, Xie, Yang, Hou, Jiao, Ren, Kong, Huang, Wu, Chen, Wang, Zhang, Wei, Li, Xu, Shen, Peng, Peng, Zhou, Li, Yang, Zhang, Xie, Huang, Lu, Fan, Cheng, Jiang, Han, Zhang, Zhu, and Ge]{huang2026step3vl10btechnicalreport}
A.~Huang, C.~Yao, C.~Han, F.~Wan, H.~Guo, H.~Lv, H.~Zhou, J.~Wang, J.~Zhou, J.~Sun, J.~Hu, K.~Lin, L.~Zhao, M.~Huang, S.~Yuan, W.~Qu, X.~Wang, Y.~Lai, Y.~Zhao, Y.~Zhang, Y.~Shi, Y.~Chen, Z.~Weng, Z.~Meng, A.~Li, A.~Kong, B.~Dong, C.~Wan, D.~Wang, D.~Qi, D.~Li, E.~Yu, G.~Li, H.~Yin, H.~Zhou, H.~Zhang, H.~Yan, H.~Zhou, H.~Peng, J.~Zhang, J.~Lv, J.~Fu, J.~Cheng, J.~Zhou, J.~Yin, J.~Xie, J.~Wu, J.~Zhang, J.~Liu, K.~Tan, K.~Yan, L.~Chen, L.~Chen, M.~Li, Q.~Zhao, Q.~Sun, S.~Pang, S.~Fan, S.~Shang, S.~Zhang, T.~You, W.~Ji, W.~Xie, X.~Yang, X.~Hou, X.~Jiao, X.~Ren, X.~Kong, X.~Huang, X.~Wu, X.~Chen, X.~Wang, X.~Zhang, Y.~Wei, Y.~Li, Y.~Xu, Y.~Shen, Y.~Peng, Y.~Peng, Y.~Zhou, Y.~Li, Y.~Yang, Y.~Zhang, Z.~Xie, Z.~Huang, Z.~Lu, Z.~Fan, Z.~Cheng, D.~Jiang, Q.~Han, X.~Zhang, Y.~Zhu, and Z.~Ge.
\newblock Step3-vl-10b technical report, 2026.
\newblock URL \url{https://arxiv.org/abs/2601.09668}.

\bibitem[Jones et~al.(2025)Jones, Mees, Sferrazza, Stachowicz, Abbeel, and Levine]{jones2025fuse}
J.~Jones, O.~Mees, C.~Sferrazza, K.~Stachowicz, P.~Abbeel, and S.~Levine.
\newblock Beyond sight: Finetuning generalist robot policies with heterogeneous sensors via language grounding.
\newblock \emph{arXiv preprint arXiv:2501.04693}, 2025.

\bibitem[Kaggle()]{fcs}
Kaggle.
\newblock Fsc dataset.
\newblock URL \url{https://www.kaggle.com/datasets/xuncngng/fsc147-0}.

\bibitem[Kim et~al.(2022)Kim, Hong, Yim, Nam, Park, Yim, Hwang, Yun, Han, and Park]{kim2022synthdog}
G.~Kim, T.~Hong, M.~Yim, J.~Nam, J.~Park, W.~Yim, S.~Hwang, S.~Yun, D.~Han, and S.~Park.
\newblock Ocr-free document understanding transformer.
\newblock In \emph{European Conference on Computer Vision (ECCV)}, 2022.

\bibitem[Kim et~al.(2024)Kim, Pertsch, Karamcheti, Xiao, Balakrishna, Nair, Rafailov, Foster, Lam, Sanketi, Vuong, Kollar, Burchfiel, Tedrake, Sadigh, Levine, Liang, and Finn]{kim24openvla}
M.~Kim, K.~Pertsch, S.~Karamcheti, T.~Xiao, A.~Balakrishna, S.~Nair, R.~Rafailov, E.~Foster, G.~Lam, P.~Sanketi, Q.~Vuong, T.~Kollar, B.~Burchfiel, R.~Tedrake, D.~Sadigh, S.~Levine, P.~Liang, and C.~Finn.
\newblock Openvla: An open-source vision-language-action model.
\newblock \emph{arXiv preprint arXiv:2406.09246}, 2024.

\bibitem[Kuznetsova et~al.(2020)Kuznetsova, Rom, Alldrin, Uijlings, Krasin, Pont-Tuset, Kamali, Popov, Malloci, Kolesnikov, et~al.]{kuznetsova2020open}
A.~Kuznetsova, H.~Rom, N.~Alldrin, J.~Uijlings, I.~Krasin, J.~Pont-Tuset, S.~Kamali, S.~Popov, M.~Malloci, A.~Kolesnikov, et~al.
\newblock The open images dataset v4: Unified image classification, object detection, and visual relationship detection at scale.
\newblock \emph{International journal of computer vision}, 128\penalty0 (7):\penalty0 1956--1981, 2020.

\bibitem[Lauren{\c{c}}on et~al.(2024)Lauren{\c{c}}on, Tronchon, and Sanh]{websight}
H.~Lauren{\c{c}}on, L.~Tronchon, and V.~Sanh.
\newblock Unlocking the conversion of web screenshots into html code with the websight dataset.
\newblock \emph{arXiv preprint}, 2024.

\bibitem[Li et~al.(2022)Li, Li, Xiong, and Hoi]{li2022blip}
J.~Li, D.~Li, C.~Xiong, and S.~Hoi.
\newblock Blip: Bootstrapping language-image pre-training for unified vision-language understanding and generation.
\newblock In \emph{International conference on machine learning}, pages 12888--12900. PMLR, 2022.

\bibitem[Li et~al.(2024)Li, Liang, Wang, Luo, Chen, Liao, Wei, Deng, Xu, Zhang, et~al.]{li2024cogact}
Q.~Li, Y.~Liang, Z.~Wang, L.~Luo, X.~Chen, M.~Liao, F.~Wei, Y.~Deng, S.~Xu, Y.~Zhang, et~al.
\newblock Cogact: A foundational vision-language-action model for synergizing cognition and action in robotic manipulation.
\newblock \emph{arXiv preprint arXiv:2411.19650}, 2024.

\bibitem[Lin et~al.(2014)Lin, Maire, Belongie, Hays, Perona, Ramanan, Doll{\'a}r, and Zitnick]{lin2014microsoft}
T.-Y. Lin, M.~Maire, S.~Belongie, J.~Hays, P.~Perona, D.~Ramanan, P.~Doll{\'a}r, and C.~L. Zitnick.
\newblock Microsoft coco: Common objects in context.
\newblock In \emph{European conference on computer vision}, pages 740--755. Springer, 2014.

\bibitem[Lipman et~al.(2022)Lipman, Chen, Ben-Hamu, Nickel, and Le]{lipman2022flowmatching}
Y.~Lipman, R.~T.~Q. Chen, H.~Ben-Hamu, M.~Nickel, and M.~Le.
\newblock Flow matching for generative modeling.
\newblock \emph{arXiv preprint arXiv:2210.02747}, 2022.
\newblock URL \url{https://arxiv.org/abs/2210.02747}.

\bibitem[Liu et~al.(2023)Liu, Zhu, Gao, Feng, Liu, Zhu, and Stone]{liu2023libero}
B.~Liu, Y.~Zhu, C.~Gao, Y.~Feng, Q.~Liu, Y.~Zhu, and P.~Stone.
\newblock Libero: Benchmarking knowledge transfer for lifelong robot learning.
\newblock \emph{arXiv preprint arXiv:2306.03310}, 2023.

\bibitem[Liu et~al.(2024)Liu, Wei, Chen, Kong, Ge, Zhu, Zhao, Sun, Han, and Zhang]{liu2024focus}
C.~Liu, H.~Wei, J.~Chen, L.~Kong, Z.~Ge, Z.~Zhu, L.~Zhao, J.~Sun, C.~Han, and X.~Zhang.
\newblock Focus anywhere for fine-grained multi-page document understanding.
\newblock \emph{arXiv preprint arXiv:2405.14295}, 2024.

\bibitem[Liu et~al.(2025)Liu, Wang, Yuan, Huang, Liu, He, and Tu]{liu2025conflict}
X.~Liu, W.~Wang, Y.~Yuan, J.~Huang, Q.~Liu, P.~He, and Z.~Tu.
\newblock Insight over sight: Exploring the vision-knowledge conflicts in multimodal llms.
\newblock \emph{arXiv preprint arXiv:2410.08145}, 2025.

\bibitem[Loshchilov and Hutter(2017)]{loshchilov2017adamw}
I.~Loshchilov and F.~Hutter.
\newblock Decoupled weight decay regularization.
\newblock \emph{arXiv preprint arXiv:1711.05101}, 2017.

\bibitem[NVIDIA et~al.(2025)NVIDIA, Bjorck, Fernando~Castañeda, Da, Ding, Fan, Fang, Fox, Hu, Huang, Jang, Jiang, Kautz, Kundalia, Lao, Li, Lin, Lin, Liu, Llontop, Magne, Mandlekar, Narayan, Nasiriany, Reed, Tan, Wang, Wang, Wang, Wang, Xiang, Xie, Xu, Xu, Ye, Yu, Zhang, Zhang, Zhao, Zheng, and Zhu]{gr00tn1_2025}
NVIDIA, J.~Bjorck, N.~C. Fernando~Castañeda, X.~Da, R.~Ding, L.~J. Fan, Y.~Fang, D.~Fox, F.~Hu, S.~Huang, J.~Jang, Z.~Jiang, J.~Kautz, K.~Kundalia, L.~Lao, Z.~Li, Z.~Lin, K.~Lin, G.~Liu, E.~Llontop, L.~Magne, A.~Mandlekar, A.~Narayan, S.~Nasiriany, S.~Reed, Y.~L. Tan, G.~Wang, Z.~Wang, J.~Wang, Q.~Wang, J.~Xiang, Y.~Xie, Y.~Xu, Z.~Xu, S.~Ye, Z.~Yu, A.~Zhang, H.~Zhang, Y.~Zhao, R.~Zheng, and Y.~Zhu.
\newblock {GR00T} {N1}: An open foundation model for generalist humanoid robots.
\newblock In \emph{ArXiv Preprint}, March 2025.

\bibitem[SakiRinn()]{locount}
SakiRinn.
\newblock Locount.
\newblock URL \url{https://github.com/SakiRinn/mmdetection-locount}.

\bibitem[Schuhmann et~al.(2022)Schuhmann, Beaumont, Vencu, Gordon, Wightman, Cherti, Coombes, Katta, Mullis, Wortsman, et~al.]{schuhmann2022laion}
C.~Schuhmann, R.~Beaumont, R.~Vencu, C.~Gordon, R.~Wightman, M.~Cherti, T.~Coombes, A.~Katta, C.~Mullis, M.~Wortsman, et~al.
\newblock Laion-5b: An open large-scale dataset for training next generation image-text models.
\newblock \emph{Advances in neural information processing systems}, 35:\penalty0 25278--25294, 2022.

\bibitem[Shi et~al.(2025)Shi, Xie, Liu, Sun, Liu, Wang, Zhou, Fan, Zhang, and Huang]{shi2025memoryvla}
H.~Shi, B.~Xie, Y.~Liu, L.~Sun, F.~Liu, T.~Wang, E.~Zhou, H.~Fan, X.~Zhang, and G.~Huang.
\newblock Memoryvla: Perceptual-cognitive memory in vision-language-action models for robotic manipulation.
\newblock \emph{arXiv preprint arXiv:2508.19236}, 2025.

\bibitem[{Sujet AI}(2024)]{sujetfinanceqavision100k}
{Sujet AI}.
\newblock Sujet-finance-qa-vision-100k: A large-scale dataset for financial document vqa, 2024.
\newblock URL \url{https://huggingface.co/datasets/sujet-ai/Sujet-Finance-QA-Vision-100k}.

\bibitem[Sun et~al.(2025)Sun, Xie, Liu, Shi, Wang, and Cao]{sun2025geovla}
L.~Sun, B.~Xie, Y.~Liu, H.~Shi, T.~Wang, and J.~Cao.
\newblock Geovla: Empowering 3d representations in vision-language-action models.
\newblock \emph{arXiv preprint arXiv:2508.09071}, 2025.

\bibitem[Team(2025{\natexlab{a}})]{galaxea2025}
G.~Team.
\newblock Galaxea g0: Open-world dataset and dual-system vla model.
\newblock \emph{arXiv preprint arXiv:2509.00576v1}, 2025{\natexlab{a}}.

\bibitem[Team et~al.(2025)Team, Kamath, Ferret, Pathak, Vieillard, Merhej, Perrin, Matejovicova, Ramé, Rivière, Rouillard, Mesnard, Cideron, bastien Grill, Ramos, Yvinec, Casbon, Pot, Penchev, Liu, Visin, Kenealy, Beyer, Zhai, Tsitsulin, Busa-Fekete, Feng, Sachdeva, Coleman, Gao, Mustafa, Barr, Parisotto, Tian, Eyal, Cherry, Peter, Sinopalnikov, Bhupatiraju, Agarwal, Kazemi, Malkin, Kumar, Vilar, Brusilovsky, Luo, Steiner, Friesen, Sharma, Sharma, Gilady, Goedeckemeyer, Saade, Feng, Kolesnikov, Bendebury, Abdagic, Vadi, György, Pinto, Das, Bapna, Miech, Yang, Paterson, Shenoy, Chakrabarti, Piot, Wu, Shahriari, Petrini, Chen, Lan, Choquette-Choo, Carey, Brick, Deutsch, Eisenbud, Cattle, Cheng, Paparas, Sreepathihalli, Reid, Tran, Zelle, Noland, Huizenga, Kharitonov, Liu, Amirkhanyan, Cameron, Hashemi, Klimczak-Plucińska, Singh, Mehta, Lehri, Hazimeh, Ballantyne, Szpektor, Nardini, Pouget-Abadie, Chan, Stanton, Wieting, Lai, Orbay, Fernandez, Newlan, yeong Ji, Singh, Black, Yu, Hui, Vodrahalli, Greff, Qiu,
  Valentine, Coelho, Ritter, Hoffman, Watson, Chaturvedi, Moynihan, Ma, Babar, Noy, Byrd, Roy, Momchev, Chauhan, Sachdeva, Bunyan, Botarda, Caron, Rubenstein, Culliton, Schmid, Sessa, Xu, Stanczyk, Tafti, Shivanna, Wu, Pan, Rokni, Willoughby, Vallu, Mullins, Jerome, Smoot, Girgin, Iqbal, Reddy, Sheth, Põder, Bhatnagar, Panyam, Eiger, Zhang, Liu, Yacovone, Liechty, Kalra, Evci, Misra, Roseberry, Feinberg, Kolesnikov, Han, Kwon, Chen, Chow, Zhu, Wei, Egyed, Cotruta, Giang, Kirk, Rao, Black, Babar, Lo, Moreira, Martins, Sanseviero, Gonzalez, Gleicher, Warkentin, Mirrokni, Senter, Collins, Barral, Ghahramani, Hadsell, Matias, Sculley, Petrov, Fiedel, Shazeer, Vinyals, Dean, Hassabis, Kavukcuoglu, Farabet, Buchatskaya, Alayrac, Anil, Dmitry, Lepikhin, Borgeaud, Bachem, Joulin, Andreev, Hardin, Dadashi, and Hussenot]{gemmateam2025gemma3technicalreport}
G.~Team, A.~Kamath, J.~Ferret, S.~Pathak, N.~Vieillard, R.~Merhej, S.~Perrin, T.~Matejovicova, A.~Ramé, M.~Rivière, L.~Rouillard, T.~Mesnard, G.~Cideron, J.~bastien Grill, S.~Ramos, E.~Yvinec, M.~Casbon, E.~Pot, I.~Penchev, G.~Liu, F.~Visin, K.~Kenealy, L.~Beyer, X.~Zhai, A.~Tsitsulin, R.~Busa-Fekete, A.~Feng, N.~Sachdeva, B.~Coleman, Y.~Gao, B.~Mustafa, I.~Barr, E.~Parisotto, D.~Tian, M.~Eyal, C.~Cherry, J.-T. Peter, D.~Sinopalnikov, S.~Bhupatiraju, R.~Agarwal, M.~Kazemi, D.~Malkin, R.~Kumar, D.~Vilar, I.~Brusilovsky, J.~Luo, A.~Steiner, A.~Friesen, A.~Sharma, A.~Sharma, A.~M. Gilady, A.~Goedeckemeyer, A.~Saade, A.~Feng, A.~Kolesnikov, A.~Bendebury, A.~Abdagic, A.~Vadi, A.~György, A.~S. Pinto, A.~Das, A.~Bapna, A.~Miech, A.~Yang, A.~Paterson, A.~Shenoy, A.~Chakrabarti, B.~Piot, B.~Wu, B.~Shahriari, B.~Petrini, C.~Chen, C.~L. Lan, C.~A. Choquette-Choo, C.~Carey, C.~Brick, D.~Deutsch, D.~Eisenbud, D.~Cattle, D.~Cheng, D.~Paparas, D.~S. Sreepathihalli, D.~Reid, D.~Tran, D.~Zelle, E.~Noland, E.~Huizenga,
  E.~Kharitonov, F.~Liu, G.~Amirkhanyan, G.~Cameron, H.~Hashemi, H.~Klimczak-Plucińska, H.~Singh, H.~Mehta, H.~T. Lehri, H.~Hazimeh, I.~Ballantyne, I.~Szpektor, I.~Nardini, J.~Pouget-Abadie, J.~Chan, J.~Stanton, J.~Wieting, J.~Lai, J.~Orbay, J.~Fernandez, J.~Newlan, J.~yeong Ji, J.~Singh, K.~Black, K.~Yu, K.~Hui, K.~Vodrahalli, K.~Greff, L.~Qiu, M.~Valentine, M.~Coelho, M.~Ritter, M.~Hoffman, M.~Watson, M.~Chaturvedi, M.~Moynihan, M.~Ma, N.~Babar, N.~Noy, N.~Byrd, N.~Roy, N.~Momchev, N.~Chauhan, N.~Sachdeva, O.~Bunyan, P.~Botarda, P.~Caron, P.~K. Rubenstein, P.~Culliton, P.~Schmid, P.~G. Sessa, P.~Xu, P.~Stanczyk, P.~Tafti, R.~Shivanna, R.~Wu, R.~Pan, R.~Rokni, R.~Willoughby, R.~Vallu, R.~Mullins, S.~Jerome, S.~Smoot, S.~Girgin, S.~Iqbal, S.~Reddy, S.~Sheth, S.~Põder, S.~Bhatnagar, S.~R. Panyam, S.~Eiger, S.~Zhang, T.~Liu, T.~Yacovone, T.~Liechty, U.~Kalra, U.~Evci, V.~Misra, V.~Roseberry, V.~Feinberg, V.~Kolesnikov, W.~Han, W.~Kwon, X.~Chen, Y.~Chow, Y.~Zhu, Z.~Wei, Z.~Egyed, V.~Cotruta, M.~Giang, P.~Kirk,
  A.~Rao, K.~Black, N.~Babar, J.~Lo, E.~Moreira, L.~G. Martins, O.~Sanseviero, L.~Gonzalez, Z.~Gleicher, T.~Warkentin, V.~Mirrokni, E.~Senter, E.~Collins, J.~Barral, Z.~Ghahramani, R.~Hadsell, Y.~Matias, D.~Sculley, S.~Petrov, N.~Fiedel, N.~Shazeer, O.~Vinyals, J.~Dean, D.~Hassabis, K.~Kavukcuoglu, C.~Farabet, E.~Buchatskaya, J.-B. Alayrac, R.~Anil, Dmitry, Lepikhin, S.~Borgeaud, O.~Bachem, A.~Joulin, A.~Andreev, C.~Hardin, R.~Dadashi, and L.~Hussenot.
\newblock Gemma 3 technical report, 2025.

\bibitem[Team(2025{\natexlab{b}})]{robochallenge}
R.~Team.
\newblock Robochallenge: Large-scale real-robot evaluation of embodied policies, 2025{\natexlab{b}}.
\newblock URL \url{https://robochallenge.cn/robochallenge_techreport.pdf}.

\bibitem[Team(2026)]{spiritai2026spiritv15}
S.~A. Team.
\newblock Spirit-v1.5: Clean data is the enemy of great robot foundation models.
\newblock \emph{Spirit AI Blog}, 2026.
\newblock https://www.spirit-ai.com/en/blog/spirit-v1-5.

\bibitem[Tong et~al.(2024)Tong, Brown, Wu, Woo, Middepogu, Akula, Yang, Yang, Iyer, Pan, Wang, Fergus, LeCun, and Xie]{tong2024cambrian1}
S.~Tong, E.~Brown, P.~Wu, S.~Woo, M.~Middepogu, S.~C. Akula, J.~Yang, S.~Yang, A.~Iyer, X.~Pan, A.~Wang, R.~Fergus, Y.~LeCun, and S.~Xie.
\newblock Cambrian-1: A fully open, vision-centric exploration of multimodal llms, 2024.

\bibitem[Touvron et~al.(2023)Touvron, Lavril, Izacard, Martinet, Lachaux, Lacroix, Rozi{\`e}re, Goyal, Hambro, Azhar, et~al.]{touvron2023llama}
H.~Touvron, T.~Lavril, G.~Izacard, X.~Martinet, M.-A. Lachaux, T.~Lacroix, B.~Rozi{\`e}re, N.~Goyal, E.~Hambro, F.~Azhar, et~al.
\newblock Llama: Open and efficient foundation language models.
\newblock \emph{arXiv preprint arXiv:2302.13971}, 2023.

\bibitem[Vuong et~al.(2023)Vuong, Levine, Walke, Pertsch, Singh, Doshi, Xu, Luo, Tan, Shah, et~al.]{vuong2023open}
Q.~Vuong, S.~Levine, H.~R. Walke, K.~Pertsch, A.~Singh, R.~Doshi, C.~Xu, J.~Luo, L.~Tan, D.~Shah, et~al.
\newblock Open x-embodiment: Robotic learning datasets and rt-x models.
\newblock In \emph{Towards Generalist Robots: Learning Paradigms for Scalable Skill Acquisition@ CoRL2023}, 2023.

\bibitem[Wang et~al.(2024)Wang, Xu, Zhao, Ouyang, Wu, Zhao, Xu, Liu, Qu, Shang, et~al.]{wang24mineru}
B.~Wang, C.~Xu, X.~Zhao, L.~Ouyang, F.~Wu, Z.~Zhao, R.~Xu, K.~Liu, Y.~Qu, F.~Shang, et~al.
\newblock Mineru: An open-source solution for precise document content extraction.
\newblock \emph{arXiv preprint arXiv:2409.18839}, 2024.

\bibitem[Wen et~al.(2025)Wen, Li, Gu, Zhao, Wang, and Sun]{wen2025llada}
Y.~Wen, H.~Li, K.~Gu, Y.~Zhao, T.~Wang, and X.~Sun.
\newblock Llada-vla: Vision language diffusion action models.
\newblock \emph{arXiv preprint arXiv:2509.06932}, 2025.

\bibitem[Wu et~al.(2025)Wu, Hou, Liu, Che, Ju, Yang, Li, Zhao, Xu, Yang, et~al.]{wu2025robomind}
K.~Wu, C.~Hou, J.~Liu, Z.~Che, X.~Ju, Z.~Yang, M.~Li, Y.~Zhao, Z.~Xu, G.~Yang, et~al.
\newblock Robomind: Benchmark on multi-embodiment intelligence normative data for robot manipulation.
\newblock In \emph{Robotics: Science and Systems (RSS) 2025}. Robotics: Science and Systems Foundation, 2025.
\newblock URL \url{https://www.roboticsproceedings.org/rss21/p152.pdf}.

\bibitem[Xie et~al.(2025)Xie, Zhou, Jia, Shi, Fan, Zhang, Li, Sun, Bin, Huang, et~al.]{xie2025dexbotic}
B.~Xie, E.~Zhou, F.~Jia, H.~Shi, H.~Fan, H.~Zhang, H.~Li, J.~Sun, J.~Bin, J.~Huang, et~al.
\newblock Dexbotic: Open-source vision-language-action toolbox.
\newblock \emph{arXiv preprint arXiv:2510.23511}, 2025.

\bibitem[Xie et~al.(2023)Xie, Cai, Li, Kong, Wu, Song, Morimitsu, Yao, Wang, Zhang, et~al.]{xie2023ccmb}
C.~Xie, H.~Cai, J.~Li, F.~Kong, X.~Wu, J.~Song, H.~Morimitsu, L.~Yao, D.~Wang, X.~Zhang, et~al.
\newblock Ccmb: A large-scale chinese cross-modal benchmark.
\newblock In \emph{Proceedings of the 31st ACM International Conference on Multimedia}, pages 4219--4227, 2023.

\bibitem[Yan et~al.(2025)Yan, Wang, Huang, Shen, Meng, Fan, Tan, Gao, Shi, Yang, et~al.]{yan2025step}
H.~Yan, J.~Wang, X.~Huang, Y.~Shen, Z.~Meng, Z.~Fan, K.~Tan, J.~Gao, L.~Shi, M.~Yang, et~al.
\newblock Step-gui technical report.
\newblock \emph{arXiv preprint arXiv:2512.15431}, 2025.

\bibitem[Yang et~al.(2025{\natexlab{a}})Yang, Li, Yang, Zhang, Hui, Zheng, Yu, Gao, Huang, Lv, Zheng, Liu, Zhou, Huang, Hu, Ge, Wei, Lin, Tang, Zhang, Tu, Zhang, Yang, Yang, Zhou, Zhou, Lin, Dang, Bao, Yang, Yu, Deng, Li, Xue, Li, Zhang, Wang, Zhu, Men, Gao, Liu, Luo, Li, Tang, Yin, Ren, Wang, Zhang, Ren, Fan, Su, Zhang, Zhang, Wan, Liu, Wang, Cui, Zhang, Zhou, and Qiu]{qwen3technicalreport2025}
A.~Yang, A.~Li, B.~Yang, B.~Zhang, B.~Hui, B.~Zheng, B.~Yu, C.~Gao, C.~Huang, C.~Lv, C.~Zheng, D.~Liu, F.~Zhou, F.~Huang, F.~Hu, H.~Ge, H.~Wei, H.~Lin, J.~Tang, J.~Zhang, J.~Tu, J.~Zhang, J.~Yang, J.~Yang, J.~Zhou, J.~Zhou, J.~Lin, K.~Dang, K.~Bao, K.~Yang, L.~Yu, L.~Deng, M.~Li, M.~Xue, M.~Li, P.~Zhang, P.~Wang, Q.~Zhu, R.~Men, R.~Gao, S.~Liu, S.~Luo, T.~Li, T.~Tang, W.~Yin, X.~Ren, X.~Wang, X.~Zhang, X.~Ren, Y.~Fan, Y.~Su, Y.~Zhang, Y.~Zhang, Y.~Wan, Y.~Liu, Z.~Wang, Z.~Cui, Z.~Zhang, Z.~Zhou, and Z.~Qiu.
\newblock Qwen3 technical report.
\newblock \emph{arXiv preprint arXiv:2505.09388}, 2025{\natexlab{a}}.
\newblock URL \url{https://arxiv.org/abs/2505.09388}.

\bibitem[Yang et~al.(2025{\natexlab{b}})Yang, Patel, Deitke, Gupta, Weihs, Head, Yatskar, Callison-Burch, Krishna, Kembhavi, and Clark]{yang2025cosyn}
Y.~Yang, A.~Patel, M.~Deitke, T.~Gupta, L.~Weihs, A.~Head, M.~Yatskar, C.~Callison-Burch, R.~Krishna, A.~Kembhavi, and C.~Clark.
\newblock Scaling text-rich image understanding via code-guided synthetic multimodal data generation.
\newblock \emph{arXiv preprint arXiv:2502.14846}, 2025{\natexlab{b}}.

\bibitem[Yu et~al.(2024)Yu, Zhao, Wei, Yang, Wu, Kong, Wei, Wang, Ge, Zhang, et~al.]{yu2024merlin}
E.~Yu, L.~Zhao, Y.~Wei, J.~Yang, D.~Wu, L.~Kong, H.~Wei, T.~Wang, Z.~Ge, X.~Zhang, et~al.
\newblock Merlin: Empowering multimodal llms with foresight minds.
\newblock In \emph{European Conference on Computer Vision}, pages 425--443. Springer, 2024.

\bibitem[Yu et~al.(2025)Yu, Lin, Zhao, Wei, Zhu, Wei, Sun, Ge, Zhang, Wang, et~al.]{yu2025unhackable}
E.~Yu, K.~Lin, L.~Zhao, Y.~Wei, Z.~Zhu, H.~Wei, J.~Sun, Z.~Ge, X.~Zhang, T.~Wang, et~al.
\newblock Unhackable temporal rewarding for scalable video mllms.
\newblock \emph{arXiv preprint arXiv:2502.12081}, 2025.

\bibitem[Yuan et~al.(2022)Yuan, Liu, Dikubab, Liu, Ji, Wu, and Bai]{hme100k}
Y.~Yuan, X.~Liu, W.~Dikubab, H.~Liu, Z.~Ji, Z.~Wu, and X.~Bai.
\newblock Syntax-aware network for handwritten mathematical expression recognition.
\newblock \emph{arXiv preprint arXiv:2203.01601}, 2022.

\bibitem[Zellers et~al.(2019)Zellers, Bisk, Farhadi, and Choi]{zellers2019vcr}
R.~Zellers, Y.~Bisk, A.~Farhadi, and Y.~Choi.
\newblock From recognition to cognition: Visual commonsense reasoning.
\newblock \emph{arXiv preprint arXiv:1811.10830}, 2019.

\bibitem[Zhang et~al.(2026)Zhang, Wang, Gao, Su, Dai, Zhou, Lu, and Tang]{zhang2026clap}
C.~Zhang, J.~Wang, Z.~Gao, Y.~Su, T.~Dai, C.~Zhou, J.~Lu, and Y.~Tang.
\newblock Clap: Contrastive latent action pretraining for learning vision-language-action models from human videos.
\newblock \emph{arXiv preprint arXiv:2601.04061}, 2026.
\newblock URL \url{https://arxiv.org/abs/2601.04061}.

\bibitem[Zitkovich et~al.(2023)Zitkovich, Yu, Xu, Xu, Xiao, Xia, Wu, Wohlhart, Welker, Wahid, et~al.]{zitkovich2023rt-2}
B.~Zitkovich, T.~Yu, S.~Xu, P.~Xu, T.~Xiao, F.~Xia, J.~Wu, P.~Wohlhart, S.~Welker, A.~Wahid, et~al.
\newblock Rt-2: Vision-language-action models transfer web knowledge to robotic control.
\newblock In \emph{Conference on Robot Learning}, pages 2165--2183. PMLR, 2023.

\end{thebibliography}

\newpage
\section{Author List}

All authors are listed alphabetically by their first names.

\paragraph{Pretraining \& Mid-training \& Post-training:} 
En Yu, Haoran Lv, Jianjian Sun, Kangheng Lin, Ruitao Zhang, Yukang Shi, Yuyang Chen, Ze Chen, Ziheng Zhang

\paragraph{Supervised Fine-tuning:} Fan Jia, Kaixin Liu, Meng Zhang, Ruitao Hao, Saike Huang, Songhan Xie, Yu Liu, Zhao Wu

\paragraph{Reinforcement Learning:} Bin Xie, Pengwei Zhang, Qi Yang, Xianchi Deng, Yunfei Wei

\paragraph{Data Collection \& Process:} Enwen Zhang, Hongyang Peng, Jie Zhao, Kai Liu, Wei Sun, Yajun Wei, Yi Yang, Yunqiao Zhang, Ziwei Yan

\paragraph{Evaluation:} Enwen Zhang, Haitao Yang, Hao Liu, Haoqiang Fan, Haowei Zhang, Junwen Huang, Kai Liu, Yang Chen, Yunchao Ma, Yunhuan Yang, Zhengyuan Du, Ziming Liu

\paragraph{Infrastructure:} Jiahui Niu, Yucheng Zhao 

\paragraph{Sponsors:} Daxin Jiang, Wenbin Tang, Xiangyu Zhang, Zheng Ge

\paragraph{Project Lead:} Erjin Zhou, Tiancai Wang

\setcounter{figure}{0}
\makeatletter
\renewcommand{\thefigure}{A\@arabic\c@figure}
\makeatother

\setcounter{table}{0}
\makeatletter
\renewcommand{\thetable}{A\@arabic\c@table}
\makeatother

\end{document}